\documentclass[%
   , hidempi
]{mpi2015-cscpreprint}
\usepackage{geometry}
\usepackage{mymacros}

\usepackage{tcolorbox}
\usepackage[tikz]{bclogo}

\usepackage{pgfplots} 
\usetikzlibrary{patterns}

\usepackage{algorithm} 
\usepackage{algpseudocode}
\usepackage{caption}
\usepackage{subcaption}
\usepackage{bbm}
\usepackage{todonotes}
\usepackage[normalem]{ulem}

\newtheorem{remark}{Remark}[section]

\numberwithin{equation}{section}
\numberwithin{theorem}{section}

\newenvironment{brsm}{% % short for 'bracketed small matrix'
	\left[ \begin{smallmatrix} }{%
	\end{smallmatrix} \right]
}
\newcommand{\noreg}{\texttt{benchmark}}
\newcommand{\nnreg}{\texttt{rRSMI(eq\_wts)}}
\newcommand{\wnnreg}{\texttt{rRSMI}}

\newcommand{\pathfig}{Results}

\renewcommand{\hat}[1]{\widehat{#1}}
\renewcommand{\tilde}[1]{\widetilde{#1}}

\definecolor{orange}{HTML}{FF7F00}
\definecolor{matplotlibcolor1}{HTML}{1f77b4}
\definecolor{matplotlibcolor2}{HTML}{ff7f0e}
\definecolor{matplotlibcolor3}{HTML}{2ca02c}
\definecolor{matplotlibcolor4}{HTML}{d62728}
\definecolor{matplotlibcolor5}{HTML}{9467bd}

\DeclareOldFontCommand{\rm}{\normalfont\rmfamily}{\mathrm}
\DeclareOldFontCommand{\sf}{\normalfont\sffamily}{\mathsf}
\DeclareOldFontCommand{\tt}{\normalfont\ttfamily}{\mathtt}
\DeclareOldFontCommand{\bf}{\normalfont\bfseries}{\mathbf}
\DeclareOldFontCommand{\it}{\normalfont\itshape}{\mathit}
\makeatletter
\DeclareOldFontCommand{\sl}{\normalfont\slshape}{\@nomath\sl}
\DeclareOldFontCommand{\sc}{\normalfont\scshape}{\@nomath\sc}
\makeatother

\begin{document}
%\title{Stability-Guaranteed Learning for Continuous Linear Systems}
\title{Rank-Minimizing and Structured Model Inference}

\author[1]{Pawan Goyal}
\affil[1]{Max Planck Institute of Dynamics of Complex Technical Systems, Magdeburg, Germany\authorcr
	\email{goyalp@mpi-magdeburg.mpg.de}, \orcid{0000-0003-3072-7780}}

\author[2]{Benjamin Peherstorfer}
\affil[2]{Courant Institute of Mathematical Sciences, New York University, USA\authorcr
	\email{pehersto@cims.nyu.edu}, \orcid{0000-0002-1558-6775}}

\author[3]{Peter Benner}
\affil[3]{Max Planck Institute of Dynamics of Complex Technical Systems, and  \newline Otto von Guericke University, Faculty of Mathematics, Magdeburg, Germany   \authorcr
	\email{benner@mpi-magdeburg.mpg.de}, \orcid{0000-0003-3362-4103}}

\abstract{
While extracting information from data with machine learning plays an increasingly important role, physical laws and other first principles continue to provide critical insights about systems and processes of interest in science and engineering. 
This work introduces a method that infers models from data with physical insights encoded in the form of structure and that minimizes the model order so that the training data are fitted well while redundant degrees of freedom without conditions and sufficient data to fix them are automatically eliminated.
The models are formulated via solution matrices of specific instances of generalized Sylvester equations that enforce interpolation of the training data and relate the model order to the rank of the solution matrices. The proposed method numerically solves the Sylvester equations for minimal-rank solutions and so obtains models of low order. 
Numerical experiments demonstrate that the combination of structure preservation and rank minimization leads to accurate models with orders of magnitude fewer degrees of freedom than models of comparable prediction quality that are learned with structure preservation alone. 
}

\novelty{\begin{itemize}
		\item We propose a data-driven model inference from data with physical insights encoded in the form of structure. 
		\item We formulate the inference problem as finding minimal-rank matrices that solve specific instances of Sylvester equations.
		\item We discusses relaxations of the rank in the objective via the weighted nuclear norm.
		\item Numerical experiments demonstrate that the combination of structure preservation and rank minimization leads to accurate models with orders of magnitude when compared to the models that are learned with structure preservation alone. 
\end{itemize}}
\maketitle

\section{Introduction}\label{sec:Intro}
Learning models that describe the response dynamics of systems from data is an ever more important building component in science and engineering. At the same time, in many applications, there is at least some knowledge available about the physics that are described by the systems of interest \cite{doi:10.1098/rsta.2016.0153,Willcox2021}. Such physical insights often can be translated into structure, e.g., symmetries, time delays, and high-order time derivatives, which can then be imposed onto the models to ensure physically meaningful response predictions. 

In this work, we introduce a method for inferring models that preserve the structure given by physical insights and, at the same time, minimize the order of the model---the number of degrees of freedom of the model parametrization---that is necessary to fit well the training data. Critically, the order is determined during training and does not have to be specified a priori.  
Minimizing the model order means that our approach finds parsimonious models by adjusting the order during the training so that data are fitted well while redundant degrees of freedom without conditions and sufficient data to fix them are eliminated. This means that the learned models have few degrees of freedom and thus can be simulated quickly to make predictions about the underlying systems. Additionally, the learned structured models can be realized as dynamical systems that describe the states in the time domain and how they behave under control inputs, which is in contrast to black-box models that only match the input-output behavior without structure. Numerical experiments demonstrate that the combination of structure preservation and order minimization leads to accurate models with orders of magnitude fewer degrees of freedom than models of comparable prediction quality that are learned with structure preservation alone.

There is a wide range of literature on learning models of physical systems from data. First, there are methods that learn models from state observations, such as dynamic mode decomposition \cite{kutz2016dynamic,rowley2009spectral,schmid2010dynamic,tu2013dynamic}, sparsity-based methods \cite{Brunton3932,Schaeffer6634,doi:10.1098/rspa.2016.0446,doi:10.1137/16M1086637}, and operator inference \cite{peherstorfer2016data,QIAN2020132401,BENNER2020113433,ErrorOpInf,mcquarrie2020data,SWK21-HOPINF,morYildGBetal21,morBenGHal22}. Closer to the approach introduced in this work are methods from the systems and control community such as the Loewner, AAA, and vector fitting techniques in the frequency domain \cite{antoulas1986scalar,morMayA07,interpbook,gosea2018data,doi:10.1137/19M1259171,Drmac2022,morDrmGB15,morWerGG21,nakatsukasa2018aaa,doi:10.1137/21M1411081} and time domain \cite{karachalios2020bilinear,peherstorfer2017data}. Closest to our work is the approach introduced in \cite{schulze2018data}, which fits structured models by interpolating the training data. However, imposing the structure can mean that there are more degrees of freedom than conditions to fix them. The authors of \cite{schulze2018data} introduce additional constraints to close the remaining degrees of freedom, which means that the learned model is not parsimonious anymore by construction and so can require more degrees of freedom than necessary for explaining the training data. Several works aim to learn structured models by fitting model parameters via gradient-based learning methods \cite{morMliG22,schwerdtner2020structure,https://doi.org/10.48550/arxiv.2209.05101,https://doi.org/10.48550/arxiv.2209.00714}. However, these methods require that the number of parameters and, thus, the order of the models is fixed a priori, and it cannot be adapted during the training. Additionally, the success of these methods critically depends on the initial guess of the model parameters, which can be challenging to obtain and typically relies on heuristic arguments.

Our approach adapts the order of the model during training to obtain parsimonious, structured models without redundant degrees of freedom. Building on model reduction \cite{rozza2007reduced,benner2015survey,morSchVR08,interpbook}, in particular, the works \cite{morBeaG09,morBenGP19}, the key ingredient is describing models that interpolate the given response data as solutions of generalized Sylvester equations, which allows formulating the training problem as finding minimal-rank matrices that solve specific instances of Sylvester equations. We then relax the rank in the objective to the weighted nuclear norm so that the corresponding optimization problems can be solved efficiently with of-the-shelf techniques \cite{candes2009exact,fazel2002matrix,recht2010guaranteed,gu2014weighted}. It is important to note that structure such as symmetry is encoded in the learned model rather than merely being weakly enforced via penalization in the loss function. In fact, in our numerical experiments, imposing structure onto the model allows the optimizer to find even lower-order models that fit the data well than if the order of the model alone is minimized without structure preservation. In our numerical experiments, the proposed approach also outperforms %, in terms of quality, 
standard techniques from black-box machine learning in terms of prediction quality by roughly a factor of two on the same training data sets.

The manuscript is structured as follows. In \Cref{sec:Prelim}, we briefly discuss traditional model reduction techniques \cite{rozza2007reduced,benner2015survey}, in particular the interpolatory techniques proposed in \cite{morBeaG09,morBenGP19}, to construct low-order models of structured systems when high-dimensional models are available. In \Cref{sec:RSMI}, we carry over the interpolatory model reduction techniques to the data-driven setting to formulate our approach that learns minimal-order, structured models from data. Extensions of our approach to symmetric systems, systems with parameters, and multiple inputs and multiple outputs are introduced in \Cref{sec:RSMIExt}. \Cref{sec:nn_relaxation} considers computational aspects of our approach and proposes a relaxation of the rank minimization via the weighted nuclear norm to make it computationally tractable. 
In \Cref{sec:numerics}, we illustrate the efficiency of the proposed methodology on various examples, and conclusions are drawn in \Cref{sec:conclusions}.

\section{Preliminaries}\label{sec:Prelim}
We briefly review systems that are given by structured response maps, which are transfer functions in the case of dynamical systems. 

\subsection{Structured systems}
The maps that we consider, which for dynamical systems are transfer functions, have the following form for a wide class of structured systems
\begin{equation}\label{eq:struc_model}
	\bH(\bs) = \bC\left(\sum\nolimits_{i=1}^q\alpha_i(\bs) \bA_i\right)^{-1}\bB,
\end{equation}
where 
\begin{equation}\label{eq:HighDimSysMat}
	\bA_1, \dots, \bA_q \in \Rnn, \quad \bB \in \Rnm, \quad \bC\in\R^{l \times n}
\end{equation}
are system matrices,  the functions $\alpha_i: \C^d \rightarrow \C$ are assumed to be smooth meromorphic functions, and $\bs \in \C^d$ is a vector argument to these functions that can include parameters of the system. Transfer functions of structured systems such as delay systems, second-order systems, and affine parameter-dependent systems can be described in the form \eqref{eq:struc_model}. For example, for a single-delay system, $\alpha_{1}(\bs),\alpha_{2}(\bs)$, and $\alpha_{3}(\bs)$ can, respectively, be  $1, \bs$, and $e^{\tau \bs}$, where $\tau$ is a delay. If the map \eqref{eq:struc_model} represents a non-parametric transfer function, then $\bs$ is scalar and takes values on the imaginary axis. 

In the following, the transfer function $\bH(\bs)$ is strictly proper, which means that 
\[
\lim\limits_{\|\bs\|\rightarrow \infty}\bH(\bs) = 0
\]
holds. Furthermore, the functions $\alpha_i, i = 1,\ldots,q$, are linear and independent, and thus, there cannot exist a non-zero vector $\boldsymbol{f} \in \C^{q}$ for which
\[
\left[\alpha_1(\bs),\ldots,\alpha_q(\bs)\right]\boldsymbol{f} = 0, \qquad \forall \bs.
\]

\subsection{Intrusive construction of low-order structured  models}\label{sec:MOR_StructuredSys}
We briefly discuss the approach introduced in \cite{morBenGP19} for constructing low-order structured models. For ease of exposition, we focus on single-input single-output (SISO) systems so that $l = m = 1$; extensions to multiple-input multiple-output (MIMO) systems are discussed later in \Cref{sec:RSMI:MIMO}. Furthermore, the functions $\alpha_i: \C \rightarrow \C$ have a scalar argument $\bs \in \C$. The extension to a multi-variate $\bs$ and $\alpha_i: \C^d \rightarrow \C$ is just a technical extension and is discussed in \Cref{subsec:multi-variate_extension}.

The aim of the model reduction approach described in \cite{morBenGP19} is to determine a low-order model
\begin{equation}\label{eq:struc_model_LSRed}
	\hat\bH(\bs) = \hat\bC\left(\sum\nolimits_{i=1}^q\alpha_i(\bs) \hat\bA_i\right)^{-1}\hat\bB,
\end{equation}
where $\hat\bA_i\in \Rrr, i = 1, \dots, q$, $\hat\bB \in \Rmr$, and $\hat\bC \in \R^{l\times r}$. The order of the $\hat\bA_1, \dots, \hat\bA_q$ matrices is $r$ and the aim is to choose $r$ much smaller than $n$, the order of the matrices of the high-dimensional model \eqref{eq:struc_model}, while ensuring accuracy within a tolerance \texttt{tol} with respect to \eqref{eq:struc_model} via
\begin{equation}\label{eq:Prelim:HTol}
	\| \bH - \hat \bH\| \leq \texttt{tol}\,,
\end{equation}
in an appropriate norm $\|\cdot\|$. The functions $\alpha_i$ in \eqref{eq:struc_model_LSRed} are the same as in \eqref{eq:struc_model}. The matrices $\hat\bA_1, \dots, \hat\bA_q \in \Rrr,$ $\hat\bB \in \Rmr$, $\hat\bC \in \R^{l\times r}$ are constructed via Petrov-Galerkin projection with the projection matrices $\bV$ and $\bW$,
\begin{equation}\label{eq:reducedSys_Cons}
	\hat\bA_i = \bW^{\top} \bA_i \bV,\quad  \hat\bB =  \bW^{\top} \bB,\quad \hat\bC = \bC \bV. 
\end{equation}
There are many ways to construct the projection matrices $\bV$ and $\bW$; see, e.g., \cite{morBeaG09,morBenGP19,breiten2016structure}. It is important to note that constructing the matrices as in \eqref{eq:reducedSys_Cons} via projection with $\bV$ and $\bW$ is intrusive because it requires having the matrices \eqref{eq:HighDimSysMat} of the high-dimensional model available either in an assembled form or via a routine that provides the vector-matrix product.

\subsection{Intrusive interpolatory model reduction}\label{sec:Prelim:InterMOR} For simplicity of presentation, we set $m=l=1$ in this section. One widely used approach to construct $\bV$ and $\bW$ is by enforcing interpolation conditions 
\begin{equation}\label{eq:InterpCond}
	\bH(\sigma_i) = \hat\bH(\sigma_i),\qquad i = 1, \dots, r,
\end{equation}
at interpolation points $\sigma_1, \dots, \sigma_r\in \C$. 
As shown in \cite{morBeaG09}, if $\sum_{i=1}^q\alpha_i(\bs)\bA_i$ is invertible for all $\bs \in \{\sigma_1, \dots, \sigma_r\}$, then the low-order model \eqref{eq:struc_model_LSRed} with the matrices \eqref{eq:reducedSys_Cons} obtained with the projection matrices
\begin{subequations}\label{eq:ProjMatrices}
	\begin{align}
		\bV &= \begin{bmatrix}  \left(\sum_{i=1}^q\alpha_i(\sigma_1) \bA_i\right)^{-1}\bB,\ldots, \left(\sum_{i=1}^q\alpha_i(\sigma_r) \bA_i\right)^{-1}\bB \end{bmatrix},\label{eq:projectionMtx_V}\\
		\bW &= \begin{bmatrix}  \left(\sum_{i=1}^q\alpha_i(\sigma_1) \bA_i\right)^{-T}\bC,\ldots, \left(\sum_{i=1}^q\alpha_i(\sigma_r) \bA_i\right)^{-T}\bC \end{bmatrix},\label{eq:projectionMtx_W}
	\end{align}
\end{subequations}
satisfies \eqref{eq:InterpCond}, provided $\sum_{i=1}^q\alpha_i(\bs)\hat\bA_i$ is also invertible for all $\bs \in \{\sigma_1, \dots, \sigma_r\}$.
Moreover, the derivative information at the interpolation points $\sigma_1, \dots, \sigma_r$ is interpolated as well,  
\begin{equation*}
	\begin{aligned}
		\dfrac{\mathrm d}{\mathrm d\bs}\bH(\sigma_i) &= 	\dfrac{\mathrm d}{\mathrm d\bs}\hat\bH(\sigma_i),\quad i = 1, \dots, r.
	\end{aligned}
\end{equation*}
When constructing the matrices $\hat\bA_1, \dots, \hat\bA_q$ defined in~\eqref{eq:reducedSys_Cons} of the low-order model \eqref{eq:struc_model_LSRed} with the $\bV$ and $\bW$ matrices defined in \eqref{eq:ProjMatrices}, then the order $r$ of the model, i.e., the size of the matrices $\hat\bA_1, \dots, \hat\bA_q$, equals the number of interpolation conditions \eqref{eq:InterpCond}. Thus, the number of given interpolation conditions in the form of the training data determines the order of the model \eqref{eq:struc_model_LSRed} that is obtained with this process. However, there can be another model of order $r_{\text{min}} < r$ that also interpolates the data \eqref{eq:InterpCond}.  Note that \eqref{eq:InterpCond} amounts to $r \times m \times l$ conditions.
In the following, we say that the order $r_{\text{min}}$ is minimal if no matrices of a lower order than $r_{\text{min}}$ exist that lead to a model \eqref{eq:struc_model_LSRed} that satisfies the interpolation conditions \eqref{eq:InterpCond}. In practice, to reduce the order of a model while maintaining good approximate quality, once the matrices $\hat\bA_1, \dots \hat\bA_{q}$ are determined with $\bV$ and $\bW$, they are often compressed via the singular value decomposition (SVD). The compression typically leads to models that deviate from the interpolation conditions \eqref{eq:InterpCond} in favor of having lower order models that approximate the high-dimensional model in the sense of \eqref{eq:Prelim:HTol}.

The procedure to interpolatory model reduction outlined in this section is intrusive because the matrices of the high-dimensional model \eqref{eq:HighDimSysMat} are needed if the projection matrices $\bV$ and $\bW$ and the model matrices $\hat\bA_1, \dots, \hat\bA_q$ are assembled as in \eqref{eq:ProjMatrices} and \eqref{eq:reducedSys_Cons}, respectively.

%%%%%%%%%%%%%%%%%%%%%%%%%%%%%%%%%%%%%%%%%%%%%
%%%%%%%%%%%%%%%%%%%%%%%%%%%%%%%%%%%%%%%%%%%%%
%%%%%%%%%%%%%%%%%%%%%%%%%%%%%%%%%%%%%%%%%%%%%
\section{Rank-minimizing and structured model inference (RSMI)}\label{sec:RSMI}
We now introduce rank-minimizing and structured model inference (RSMI) to find a structured low-order system model with the following map
\begin{equation}\label{eq:HTilde}
	\tilde\bH(\bs) = \tilde\bC\left(\sum\nolimits_{i = 1}^q \alpha_i(\bs)\tilde\bA_i\right)^{-1}\tilde\bB
\end{equation}
from transfer function measurements 
of a system \eqref{eq:struc_model} 
\begin{equation}\label{eq:TrainingData}
	\{(\sigma_1, \bH(\sigma_1)), \dots, (\sigma_\cN, \bH(\sigma_\cN))\}
\end{equation}
and the functions $\alpha_1, \dots, \alpha_q$, but without access to the system matrices of the unknown system \eqref{eq:HighDimSysMat}. 

We build on a formulation of the projection matrices $\bV$ and $\bW$ defined in \eqref{eq:ProjMatrices} as solutions to generalized Sylvester equations, which we use to derive Sylvester equations for the matrices $\hat\bA_1, \dots, \hat\bA_q$ that depend on the training data \eqref{eq:TrainingData} only. This then motivates the objective function of the proposed RSMI approach for inferring matrices $\tilde\bA_1, \dots, \tilde\bA_q$ of minimal rank that satisfy the generalized Sylvester equations and so give rise to a model that fits the training data well while ensuring a low-order representation. 

\subsection{Model matrices as solutions of generalized Sylvester equations}
The projection matrices $\bV$ and $\bW$ in \eqref{eq:projectionMtx_V} and \eqref{eq:projectionMtx_W}, respectively, solve the following generalized Sylvester equations:
\begin{subequations}\label{eq:Slyvester}
	\begin{align}
		\bA_1\bV \Lambda_{1} + \cdots +   \bA_q\bV \Lambda_{q} &= \bB \mathbbm{1}^{\top} ,\label{eq:Slyvester_V}\\
		\bA_1^{\top} \bW \Lambda_{1} + \cdots +   \bA_q^{\top} \bW \Lambda_{q} &= \bC^{\top}  \mathbbm{1}^{\top} ,\label{eq:Slyvester_W}
	\end{align}
\end{subequations}
where $\Lambda_i = \diag{\alpha_i(\sigma_1),\ldots,\alpha_i(\sigma_\cN)}$ and $\mathbbm{1}$ is a column vector of ones. Multiplying \eqref{eq:Slyvester_V} and \eqref{eq:Slyvester_W} by the matrices $\bW^{\top} $ and $\bV^{\top} $, respectively, from the left leads to
\begin{subequations}\label{eq:red_Sly}
	\begin{align}
		\bW^{\top} \bA_1\bV \Lambda_{1} + \cdots +   \bW^{\top} \bA_q\bV \Lambda_{q} &= \bW^{\top} \bB \mathbbm{1}^{\top},\label{eq:Slyvester_V1}\\
		\bV^{\top} \bA_1^{\top} \bW \Lambda_{1} + \cdots +   \bV^{\top} \bA_q^{\top} \bW \Lambda_{q} &= \bV^{\top} \bC^{\top}  \mathbbm{1}^{\top}.\label{eq:Slyvester_W1}
	\end{align}
\end{subequations}
Now, notice that $\bW^{\top} \bA_i\bV = \hat\bA_i$ for $i = 1, \dots, q$ and
\[
\bW^{\top} \bB = \hat\bB = \begin{bmatrix}  \bH(\sigma_1), \ldots,  \bH(\sigma_\cN) \end{bmatrix}^{\top},\qquad \bC \bV= \hat\bC = \begin{bmatrix}  \bH(\sigma_1), \ldots,  \bH(\sigma_\cN) \end{bmatrix}\,.
\]
Consequently, we can write \eqref{eq:red_Sly} using only matrices that define a model:
\begin{subequations}\label{eq:red_Sly1}
	\begin{align}
		\hat\bA_1 \Lambda_{1} + \cdots +   \hat\bA_q \Lambda_{q} &=  \bH_\sigma\mathbbm{1}^{\top} ,\label{eq:Slyvester_Red1}\\
		\hat\bA_1^{\top}  \Lambda_{1} + \cdots +   \hat\bA_q^{\top}  \Lambda_{q} &= \bH_\sigma \mathbbm{1}^{\top}\,,\label{eq:Slyvester_Red2}
	\end{align}
\end{subequations}
where 
\[
\bH_\sigma = \begin{bmatrix}  \bH(\sigma_1) ,  \ldots , \bH(\sigma_\cN) \end{bmatrix}^{\top}\,.
\]
It is important to note that \eqref{eq:red_Sly1} is independent of the system matrices of the unknown model \eqref{eq:struc_model} from which data \eqref{eq:TrainingData} are sampled. Thus, the solutions of the equations \eqref{eq:red_Sly1} provide us directly with matrices \eqref{eq:reducedSys_Cons} of models that interpolate the data \eqref{eq:TrainingData}, without access to the high-dimensional matrices \eqref{eq:HighDimSysMat}. This is in contrast to finding the matrices $\hat\bA_1, \dots, \hat\bA_q$ via projection as in \eqref{eq:reducedSys_Cons}, which is based on the projection matrices $\bV$ and $\bW$ defined in \eqref{eq:ProjMatrices} that are assembled with matrix-vector products with the system matrices of the unknown model. However, directly solving equations \eqref{eq:red_Sly1} for the matrices $\hat\bA_1, \dots, \hat\bA_q$ leads to matrices that form a model of order that scales with the number of data points $\mathcal{N}$, which potentially is much higher than the minimal order required to interpolate the data; see also the comments in Section~\ref{sec:Prelim:InterMOR}.

\subsection{Uniqueness of solution of generalized Sylvester equations} All matrices $\hat\bA_1, \dots, \hat\bA_q$ that solve \eqref{eq:red_Sly1} lead to models with map $\hat\bH$ that interpolate the training data \eqref{eq:TrainingData}. If $q =  2$, it can be shown that the matrices $\hat\bA_1$ and $\hat\bA_2$ satisfy the following Sylvester equations
\begin{subequations}\label{eq:Sly_A12}
	\begin{align}
		\Lambda_{2}^{\top}  \hat\bA_1 \Lambda_{1} - \Lambda_{1}^{\top}  \hat\bA_1 \Lambda_{2}  &=  \Lambda_{2}^{\top} \bH_\sigma\mathbbm{1}^{\top}  - \mathbbm{1}\bH_\sigma ^{\top} \Lambda_2^{\top} , \\ %\label{eq:Slyvester_Red1}\\
		\Lambda_{2}^{\top}  \hat\bA_2 \Lambda_{1} -\Lambda_{1}^{\top}  \hat\bA_2 \Lambda_{2} &=  \Lambda_1^{\top}  \bH_\sigma\mathbbm{1}^{\top}  - \mathbbm{1} \bH_\sigma ^{\top} \Lambda_1^{\top}\,,
	\end{align}
\end{subequations}
which 
have a unique solution if derivative information of $\bH(\bs)$ at $\sigma_i$ is available, see \cite{morMayA07,schulze2018data}. Note that the model described by the solution matrices of \eqref{eq:Sly_A12} can be represented differently, such as via basis transformations and potentially truncation. It further is shown in \cite{schulze2018data} that the solution of \eqref{eq:Sly_A12} can be given analytically by modifying the Loewner and shifted Loewner matrices \cite{morMayA07}. For Rayleigh-damped second-order systems, the analytical solution of \eqref{eq:Sly_A12} is discussed in \cite{morDuffGB19}, and for a class of delay systems in~\cite{SchU16}. 

In the more general case, when $q\geq 2$, there are more degrees of freedom than the number of equations, and thus, there can be arbitrarily many solutions; see the discussion about the work \cite{schulze2018data} in the introduction in \Cref{sec:Intro}.

\subsection{Rank minimization and structure-preserving models}
To motivate our RSMI formulation, we first state a result from \cite{morBenGP19} about how the rank of appropriate matrices, containing $\bW^\top \bA_i\bV $, or $\hat \bA_i$ in \eqref{eq:Slyvester} which enforces interpolating conditions, relates to the order of models: Consider the training data \eqref{eq:TrainingData} and notice that the existence of $\bH(\sigma_1), \dots, \bH(\sigma_\cN)$ in the second component of the tuples in \eqref{eq:TrainingData} implies that
\[
\sum_{i = 1}^q \alpha_i(\bs)\bA_i
\]
is invertible for all $s \in \{\sigma_1, \dots, \sigma_\cN\}$.  Let now $\bV$ and $\bW$ be the projection matrices defined in \eqref{eq:ProjMatrices}, define the matrices
\begin{subequations}
	\begin{align}
		\hat{\mathcal{A}} & = \begin{bmatrix}  \bW^{\top}   \bA_1\bV,\ldots,\bW^{\top}   \bA_q\bV\end{bmatrix} =  \begin{bmatrix}  \hat\bA_1,\ldots,   \hat\bA_q\end{bmatrix}\,,
		\\
		\hat{\mathcal{A}}_T & = \begin{bmatrix}  \bW^{\top}   \bA_1\bV \\ \vdots \\ \bW^{\top}   \bA_q\bV\end{bmatrix} =  \begin{bmatrix}  \hat\bA_1 \\ \vdots\\   \hat\bA_q\end{bmatrix} \,,
	\end{align}
\end{subequations}
and set 
\begin{equation}\label{eq:rank_determine}
	r_{\text{min}} = \min\left\{\operatorname{rank}(\hat{\mathcal{A}}),\, \operatorname{rank}(\hat{\mathcal{A}}_T)\right\}\,.
\end{equation}
It is shown in \cite{morBenGP19} that there exists a model of order $r_{\text{min}}$ 
that interpolates the training data \eqref{eq:TrainingData} and that the order $r_{\text{min}}$ is minimal. Moreover, an $r_{\text{min}}$-order model can be constructed from the matrices $\hat\bA_1, \dots, \hat\bA_q$ satisfying \eqref{eq:red_Sly1} via projection onto appropriate $r_{\text{min}}$-dimensional subspaces; see \cite{morBenGP19} for more details.

The insights from the previous paragraph motivate us to look for solutions of the Sylvester equations \eqref{eq:red_Sly1} that have minimal rank, which leads to the objective
\begin{equation}\label{eq:RankObj}
	\cJ(\tilde\bA_1, \dots, \tilde\bA_q) = \min\left\{\operatorname{rank}\left([\tilde\bA_1, \dots, \tilde\bA_q]\right),\, \operatorname{rank}\left([\tilde\bA_1^{\top}, \dots, \tilde\bA_q^{\top}]^{\top}\right)\right\}\,,
\end{equation}
and the optimization problem
\begin{equation}\tag{RSMI}\label{eq:HardRankProb}
	\begin{aligned}
		\min_{\tilde\bA_1, \dots, \tilde\bA_q} ~~~~& \cJ(\tilde\bA_1, \dots, \tilde\bA_q)\\
		\text{subject to~~~~~} & ~\tilde\bA_1\Lambda_{1} + \cdots + \tilde\bA_q\Lambda_{q} =\bH_\sigma \mathbbm{1}^{\top},\\
		& \tilde\bA_1^{\top}  \Lambda_1 + \cdots + \tilde\bA_q^{\top}  \Lambda_q = \bH_\sigma \mathbbm{1}^{\top}\,,
	\end{aligned}
\end{equation}
where $\Lambda_i, i = 1, \dots, q$ and $\bH_\sigma$ are defined in \eqref{eq:Slyvester} and \eqref{eq:red_Sly1}, respectively.
Thus, in \eqref{eq:HardRankProb}, we are seeking matrices $\tilde \bA_1, \dots, \tilde \bA_q$ such that they give rise to a low-order model that interpolates the training data because we know that the objective \eqref{eq:RankObj} of \eqref{eq:HardRankProb} minimizes the rank of \eqref{eq:rank_determine}, which also determines the order of the minimal model. Consequently, once we identified $\tilde \bA_1, \dots, \tilde \bA_q$ with \eqref{eq:HardRankProb}, we can use them to construct low-order models following \cite{morBenGP19}. 

\subsection{An illustrative example} 
Consider a scalar delay example as follows:
\begin{equation}
	\begin{aligned}
		\dot \bx(t) &= -\bx(t) + 0.25\bx(t-1) + \bu(t),\\
		\by(t) &= \bx(t).
	\end{aligned}
\end{equation}
The transfer function of the above system is $\bH(\bs) = (s+1-0.25e^{-s})^{-1}$. Let us assume that we have two measurements $\bH(\sigma_1)$ and $\bH(\sigma_2)$ at $\sigma_1$ and $\sigma_2$.
Using these two data points, we would like to learn a model of the form:
\begin{equation}\label{eq:DD_Realization}
	\begin{aligned}
		\bE\bz(t) &= \bA\bz(t) + \bA_\tau \bz(t-1) + \bB\bu(t),\\
		\by(t) &=  \bC\bz(t),
	\end{aligned}
\end{equation}
ensuring 
\begin{equation}\label{eq:interpolationproperties}
	\bH_z(\sigma_i) = \bH(\sigma_i), i\in \{1,2\},
\end{equation}
where $\bH_z(\bs) = \bC\left(s\bE -\bA -e^{-s}\bA_\tau\right)^{-1}\bB$. We now know that if the matrices $\bE,\bA,\bA_\tau$ satisfy
\begin{subequations}\label{eq:demoexample}
	\begin{align}
		\bE\begin{brsm} \sigma_1 & \\ & \sigma_2 \end{brsm} + \bA\begin{brsm} 1 & \\ & 1 \end{brsm} +  \bA_\tau\begin{brsm} e^{-\sigma_1} & \\ & e^{-\sigma_2} \end{brsm} &= \begin{brsm} \bH(\sigma_1) \\ \bH(\sigma_2) \end{brsm}\mathbbm{1}^{\top} ,\\
		\bE^{\top} \begin{brsm} \sigma_1 & \\ & \sigma_2 \end{brsm} + \bA^{\top} \begin{brsm} 1 & \\ & 1 \end{brsm} +  \bA_\tau^{\top} \begin{brsm} e^{-\sigma_1} & \\ & e^{-\sigma_2} \end{brsm} &= \begin{brsm} \bH(\sigma_1) \\ \bH(\sigma_2) \end{brsm}\mathbbm{1}^{\top} ,
	\end{align}
\end{subequations}
and $\bB = \bC^{\top}  =  [ \bH(\sigma_1)^{\top},  \bH(\sigma_2)^{\top}]^{\top}$, then the model given by the matrices $\bE$, $\bA$, $\bA_\tau$, $\bB$, $\bC$ interpolates the data.  However, there clearly are arbitrarily many solutions of \eqref{eq:demoexample} for $\bE,\bA,\bA_\tau$. 

We now discuss two particular solutions. First, we set $\bA_\tau = 0$ and solve \eqref{eq:demoexample} for $\bE$ and $\bA$. It means that we approximate the time-delay system with a rational function. In this case, when we observe the rank of the matrix $[ \bE,\bA,\bA_\tau]$, it is $2$; hence, we cannot recover the original system. But it interpolates the data. In contrast, if we determine a model using the rank-based optimization problem, i.e.,
\begin{equation}
	\min_{\bE,\bA,\bA_\tau} \rank{\begin{bmatrix} \bE,\bA,\bA_\tau\end{bmatrix}}\quad \text{subject to~\eqref{eq:demoexample}},
\end{equation}
then we can get the solution
\begin{equation}
	\begin{aligned}
		\bE &= \begin{brsm} \bH(\sigma_1) & \\ & \bH(\sigma_2) \end{brsm} \begin{bmatrix} 1&1 \\ 1 &1\end{bmatrix} \begin{brsm} \bH(\sigma_1) & \\ & \bH(\sigma_2) \end{brsm},\\
		\bA &= -\begin{brsm} \bH(\sigma_1) & \\ & \bH(\sigma_2) \end{brsm}\begin{bmatrix} 1&1 \\ 1 &1\end{bmatrix}\begin{brsm} \bH(\sigma_1) & \\ & \bH(\sigma_2) \end{brsm}, \\
		\bA_\tau &= 0.25\begin{brsm} \bH(\sigma_1) & \\ & \bH(\sigma_2) \end{brsm}\begin{bmatrix} 1&1 \\ 1 &1\end{bmatrix}\begin{brsm} \bH(\sigma_1) & \\ & \bH(\sigma_2) \end{brsm}. 
	\end{aligned}
\end{equation}
It can be easily seen that $\rank{[ \bE,\bA,\bA_\tau]}=1$. After a compression step, as will be shown later in \Cref{algo:minimal_rank_realization}, we can recover a model of the original system.

\subsection{Algorithmic description of RSMI} 
We now present an algorithm to construct lower-order models by projections on the dominant left and right subspaces once the optimization problem \eqref{eq:RankObj} is solved. The algorithm is inspired from \cite[Algo. 1]{morBenGP19}. The algorithm is sketched in \Cref{algo:minimal_rank_realization}.
In the first step, we solve the optimization problem \eqref{eq:HardRankProb} to obtain $\bar \bA_1,\ldots,\bar \bA_q$. Step 2 computes singular value decompositions of appropriately constructed matrices using  $\bar \bA_1,\ldots,\bar \bA_q$. It allows us to determine a suitable order of a model that approximates well the given data in Step 3. It is followed by determining dominant subspaces $\bV$ and $\bW$ in Step 4 that allow us to construct the desired low-order model with matrices $\tilde\bA_1, \ldots, \tilde\bA_q,\tilde\bB,\tilde\bC$ in Step 6. 

\begin{algorithm}[tb]
	\caption{Obtaining low-order models via projection onto dominant subspaces.}\label{algo:minimal_rank_realization}
	\textbf{Input:} Samples $\{\sigma_i,\bH(\sigma_i)\}, i\in\{1,\ldots,\cN\}$, and functions $\alpha_i(\cdot)$.
	\begin{algorithmic}[1]
		\State Formulate and solve the rank minimization problem as in \eqref{eq:HardRankProb} and denote its solution by $\bar \bA_1,\ldots,\bar \bA_q$.
		\State Compute SVDs of the following matrices:\label{step:step2}
		\begin{align*}
			\bU_1\Sigma_1\bV_1^{\top}  &= \begin{bmatrix} \bar \bA_1,\ldots,\bar \bA_q \end{bmatrix},~~\text{and} ~~~~ \bU_2\Sigma_2\bV_2^{\top}  = \begin{bmatrix} \bar \bA_1\\\vdots\\\bar \bA_q \end{bmatrix}. 
		\end{align*}
		\State Determine the order $r$ such that
		\[
		\max\left\{\dfrac{\sum_{i=1}^r\gamma_1^{(i)}}{\sum_i\gamma_1^{(i)}}, \dfrac{\sum_{i=1}^r\gamma_2^{(i)}}{\sum_i\gamma_2^{(i)}}\right\} \leq \texttt{tol}\,,
		\]
		where $\gamma_1^{(i)}$ and $\gamma_2^{(i)}$ are the $i$-th diagonal entries of $\Sigma_1$ and $\Sigma_2$.
		\State Determine projection matrices: $\bW = \bU_1^{(r)}$ and $\bV = \bV_2^{(r)}$, where $\bU_1^{(r)}$ and $\bV_2^{(r)}$ denote the matrices that contain as columns the first $r$ dominant vectors of $\bU_1$ and $\bV_2$, respectively.
		\State Construct $\bH_\sigma = [	\bH(\sigma_1) , \ldots, \bH(\sigma_\cN) ]$.
		\State Construct the low-order model of the form \eqref{eq:struc_model} with matrices
		\begin{equation*}
			\tilde\bA_i = \bW^{\top} \bar\bA_i\bV, i = 1, \dots, q,\quad \tilde\bB = \bW^{\top} \bH_\sigma^{\top}, \quad\tilde\bC = \bH_\sigma\bV. 
		\end{equation*}
	\end{algorithmic}
	\Return $\tilde \bA_1, \ldots, \tilde\bA_q, \tilde\bB, \tilde\bC$.
\end{algorithm}

%%%%%%%%%%%%%%%%%%%%%%%%%%%%%%%%%%%%%%%%%%%%%
%%%%%%%%%%%%%%%%%%%%%%%%%%%%%%%%%%%%%%%%%%%%%
%%%%%%%%%%%%%%%%%%%%%%%%%%%%%%%%%%%%%%%%%%%%%
\section{Extensions of RSMI}\label{sec:RSMIExt} In this section, we discuss several extensions to RSMI.
\subsection{Systems with multiple inputs and multiple outputs}\label{sec:RSMI:MIMO} 
To extend RSMI to MIMO systems, we build on the concept of tangential interpolation \cite{gallivan2004model}. To that end, for given data set \eqref{eq:TrainingData} with $\bH(\sigma_i) \in \C^{l\times m}, i = 1, \dots, \cN$, now being $l \times m$ matrices, we seek to identify a structured model of the form \eqref{eq:HTilde} 
such that 
\begin{equation}\label{eq:TangInterpCond}
	\begin{aligned}
		\bH(\sigma_i)\bb_i &=  \hat\bH(\sigma_i)\bb_i\,,\qquad i = 1, \dots, \cN\,,\\
		\bc_i^{\top} \bH(\sigma_i) &=  \bc_i^{\top}\hat\bH(\sigma_i)\,,\qquad i = 1, \dots, \cN,\end{aligned}
\end{equation}
where $\bb_i \in \C^m$ and $\bc_i \in \C^l$ are right and left tangential directions, respectively, for $i = 1, \dots, \cN$. The interpolation conditions \eqref{eq:TangInterpCond} lead to generalized Sylvester equations with different right-hand sides than \eqref{eq:red_Sly1},
\begin{subequations}\label{eq:constraints_MIMO}
	\begin{align}
		\hat\bA_1\Lambda_{1} + \cdots + \hat\bA_q\Lambda_{q} &= \begin{bmatrix} \bc_1^{\top}\bH_{\sigma_1} \\ \vdots \\ \bc_\cN^{\top}\bH_{\sigma_\cN} \end{bmatrix} \cB,\\
		\hat\bA_1^{\top}  \Lambda_1 + \cdots + \hat\bA_q^{\top}  \Lambda_q &= \begin{bmatrix} \bb_1^{\top}\bH_{\sigma_1}^{\top} \\ \vdots \\ \bb_\cN^{\top}\bH_{\sigma_\cN}^{\top} \end{bmatrix} \cC,
	\end{align}
\end{subequations}
where $\cB = [ \bb_1,\ldots, \bb_\cN]$ and $\cC = [ \bc_1,\ldots, \bc_\cN]$. Equations \eqref{eq:constraints_MIMO} are then used as constraints when minimizing the objective $\cJ$ defined in \eqref{eq:RankObj}. The rest of the procedure of RSMI, for example, solving optimization problems and constructing lower-order models using dominant subspaces, remains the same as for the SISO case.

\subsection{Symmetric systems}\label{subsec:imposesymm}
The response map is symmetric if, for the data, it holds that $\bH(\bs) = \bH^{\top} (\bs)$ for all $\bs$. We then typically want to preserve that symmetry by constraining the matrices in the RSMI  optimization problem \eqref{eq:HardRankProb} to be symmetric. %In many cases, it might be desirable to obtain a symmetric realization, e.g., in mechanical applications.   
We derive the approach for the SISO case for ease of exposition, but it readily generalizes to the MIMO case following Section~\ref{sec:RSMI:MIMO}.

First, note that the constraint $\tilde\bA_i = \tilde\bA_i^{\top}$ for $i = 1, \dots, q$ means that the two constraints in \eqref{eq:HardRankProb} coincide. Furthermore, the equality
\begin{equation}
	\rank{\begin{bmatrix} \hat \bA_1,\ldots,\hat \bA_q \end{bmatrix}} = \rank{\begin{bmatrix} \hat \bA_1\\ \vdots\\\hat \bA_q \end{bmatrix}}
\end{equation}
holds. To avoid having to add the symmetry constraints explicitly as $\tilde\bA_i = \tilde\bA_i^{\top}$, we parametrize $\tilde\bA_i$ as $\tilde\bA_i = \tilde\bK_i + \tilde\bK_i^{\top}$ for $i = 1, \dots r$. We then obtain the objective
\[
\cJ_s(\tilde\bK_1, \dots, \tilde\bK_q) = \operatorname{rank}\left([\tilde\bK_1 + \tilde\bK_1^{\top}, \dots, \tilde\bK_q + \tilde\bK_q^{\top}]\right)\,.
\]
Hence, the optimization problem becomes
\begin{equation}\tag{sRSMI}\label{eq:sRSMI}
	\begin{aligned}
		\min_{\tilde\bK_1, \dots, \tilde\bK_q} ~~~~& \cJ_s(\tilde\bK_1, \dots, \tilde\bK_q)\\
		\text{subject to~~~~~} & (\tilde\bK_1 + \tilde\bK_1^{\top})\Lambda_{1} + \cdots + (\tilde\bK_q + \tilde\bK_q^{\top})\Lambda_{q} =\bH_\sigma \mathbbm{1}^{\top}\,,
	\end{aligned}
\end{equation}
with $\Lambda_1, \dots, \Lambda_q$ and $\bH_{\sigma}$ defined as in the RSMI problem \eqref{eq:HardRankProb}. From the matrix $\tilde\bK_i$, the matrix $\tilde\bA_i = \tilde\bK_i + \tilde\bK_i^{\top}$ can be readily constructed for $i = 1, \dots, q$. 
Furthermore, Step~2 in \Cref{algo:minimal_rank_realization} simplifies because $\bU_1\Sigma_1\bV_1^{\top}  = \bV_2\Sigma_2\bU_2^{\top} $, which means that $\bW = \bV$ in Step 4.

\subsection{Systems with parameters}\label{subsec:multi-variate_extension}
So far, we developed RSMI for functions $\alpha_i:\C \rightarrow \C, i = 1, \dots, q$ in \eqref{eq:struc_model} with scalar input $\bs$.  
Although the discussions so far readily cover parametric cases, it is worthwhile to explicitly highlight that we can apply RSMI for stationary parametric problems, which are of the form:
\begin{equation}
	\begin{aligned}
		\left(\alpha_1(\bp)  \bA_1 + \cdots + \alpha_q(\bp) \bA_q)\right) \bX(\bp) &= \bB, \\
		\by(\bp) &= \bC \bX(\bp). 
	\end{aligned}
\end{equation}
Hence, we can obtain structured models for parametric systems, which interpolate the outputs at given parameter values and are of low order. In our numerical section, we will illustrate this scenario using a thermal block example. 

\section{Relaxation of rank-minimization problem}\label{sec:nn_relaxation}
The objective \eqref{eq:HardRankProb} depends on the rank of a matrix. Directly solving rank-minimization problems is challenging because they lead to non-convex NP-hard optimization problems. In this section, we build on relaxations of rank-minimization problems based on the nuclear norm \cite{fazel2002matrix,fazel2001rank,recht2010guaranteed}, which is the sum of singular values.
This results in optimization problems that can be solved efficiently with gradient-based methods. However, even though the nuclear norm leads to the best convex relaxation of rank-minimization objectives, the solutions of the relaxed problems can be far from the solutions of the original problem. To address this issue, the concept of the weighted nuclear norm was proposed in \cite{gu2014weighted}, which, in practice, often yields a solution closer to the solution of the original rank-minimization problem.

\subsection{Weighted nuclear-norm formulation of RSMI}
The weighted nuclear norm of a matrix $\bT\in \Rnm$ with $m\geq n$ and weight weight $\bw \in \Rn$ is 
\begin{equation*}
	\|\bT\|_{\bw,*} = \sum_{i=1}^n w_i\gamma_i,
\end{equation*}
where $\gamma_i$ are the singular values of the matrix $\bT$ with $\gamma_{i+1}\geq \gamma_i$, and $w_i$ is the $i^{\text{th}}$ component of the vector $\bw$ \cite{gu2014weighted,mohan2012iterative}. 
We consider a vector $\bw$ with $w_i \geq w_{i-1}, i = 2, \dots, n$. Defining the vector $\bw$ with non-decreasing weights gives larger weights to smaller singular values. As a result, the smaller singular values in the course of the optimization tend to be even smaller due to a higher weight and so are nudged closer to the solution of the rank-minimization problem. The weighted nuclear norm with non-decreasing vector $\bw$ is also a convex function \cite{gu2014weighted}. Note that $\bw = [1, \dots, 1]^{\top}$ is a feasible non-decreasing weight vector, which coincides with the classical nuclear norm. 

Based on the weighted nuclear norm, we formulate the objective $\cJ_r$ as
\begin{multline}\tag{rRSMI}\label{eq:rRSMI}
	\cJ_r(\tilde\bA_1, \dots, \tilde\bA_q, \lambda) = \lambda\|[\tilde\bA_1, \dots, \tilde\bA_q] \|_{\bw,*} + \lambda\|[\tilde\bA_1^{\top}, \dots, \tilde\bA_q^{\top}]^{\top} \|_{\bw,*} + \\\|\mathcal{R}_1(\tilde\bA_1, \dots, \tilde\bA_q)\| + \|\mathcal{R}_2(\tilde\bA_1, \dots, \tilde\bA_q)\|
\end{multline}
with the terms
\begin{equation}
	\begin{aligned}
		\mathcal{R}_1(\tilde\bA_1, \dots, \tilde\bA_q) = & \tilde\bA_1\Lambda_1 + \dots + \tilde\bA_q\Lambda_q - \bH_{\sigma}\mathbbm{1}^{\top},\\
		\mathcal{R}_2(\tilde\bA_1, \dots, \tilde\bA_q) = & \tilde\bA_1^{\top}\Lambda_1 + \dots + \tilde\bA_q^{\top}\Lambda_q - \bH_{\sigma}\mathbbm{1}^{\top}\,,
	\end{aligned}
\end{equation}
that are obtained from the constraints in \eqref{eq:HardRankProb} with the Lagrange multiplier $\lambda$. 

The weight vectors influence the optimization process. Typically, a good choice for the weight vector can be based on the singular values of the solution, and the weights can be updated iteratively during the optimization. In our setting of identifying response maps such as transfer functions of systems, the singular vectors corresponding to the most significant singular values are important for capturing the system dynamics. In contrast, the singular vectors corresponding to small singular values carry little information about the system dynamics and, thus, typically can be safely ignored. We, therefore, define the weights inversely proportional to the singular values in the following. Such a scheme has been initially discussed in \cite{zhong2015nonconvex}, which is inspired by the re-weighting scheme proposed in \cite{candes2008enhancing} for weighted $l_1$ problems to obtain sparse solutions.  

\subsection{Algorithm to solve relaxed RSMI problem}
In \cref{algo:IterAlgo_WNN}, we show a computational procedure for solving the relaxed optimization problem of RSMI based on the weighted-nuclear norm subject to linear constraints. In the first step, we solve \eqref{eq:rRSMI} with a constant weight vector $\bw = [1,\ldots, 1]$ because we do not have a good choice for the weight vector at the beginning. In Steps 3 and 4, at iteration~$i$, we update the weight vector $\bw$ based on the current solution $\tilde\bA_0^{(i)}, \ldots, \tilde\bA_q^{(i)}$. For this, we compute the singular values of the matrix $[\tilde\bA_0^{(i)}, \ldots, \tilde\bA_q^{(i)}]$ and define the weight vector to have as components the inverse of the singular values plus a safety threshold for avoiding division by zero. 
With the updated weighting vector, we solve \eqref{eq:rRSMI} again and continue the iterations. We also can vary the regularization parameter with each iteration. Typically, we use regularization parameters that are non-decreasing with iterations; see numerical results in Section~\ref{sec:numerics}. Once we have a solution, then we can obtain a lower-order model using Steps 2-6 of \Cref{algo:minimal_rank_realization}. 

\begin{remark} 
	It is often desirable to obtain models with matrices with real entries only, but interpolation points $\sigma_i$ and training data $\{\sigma_i, \bH(\sigma_i)\}$ can be complex. If the data comes from a real model and training data are closed under conjugation, one can transform the inferred model into a real model as described in, e.g., \cite{schulze2018data}.
\end{remark}

\begin{algorithm}[tb]
	\caption{An iterative scheme to solve the relaxed RSMI problem.}\label{algo:IterAlgo_WNN}
	\textbf{Input:} Samples $\bH(\sigma_i), i\in\{1,\ldots,\cN\}$,  functions $\alpha_i(\cdot)$, number of iterations (\texttt{iters}), regularizing parameters $\lambda^{(0)}, \lambda^{(1)}, \dots, \lambda^{(\text{iters})}$.
	\begin{algorithmic}[1]
		\State Solve the optimization problem \eqref{eq:rRSMI} with regularizing parameter $\lambda^{(0)}$ and the constant weight vector $\bw = [1,\ldots,1]$. Denote the solution as $\tilde\bA_1^{(0)},\ldots, \tilde\bA_q^{(0)}$.
		\For{$i = 1,\ldots, \texttt{iters}$}
		\State Compute singular values $s_1, \dots, s_{\cN}$ of $[\tilde\bA_1^{(i-1)},\ldots, \tilde\bA_q^{(i-1)}]$.
		\State \parbox[t]{313pt}{Define the $j$-th component of the weight vector $\bw$ as $$\bw_j = \min\{\texttt{max\_val}, 1./(s_j+ \epsilon)\}.$$ The threshold $\epsilon$ is a small number to avoid division by zero and \texttt{max\_val} prevents too high values in the weight vector. }
		\State  \parbox[t]{313pt}{Solve \eqref{eq:rRSMI} with regularizing parameter $\lambda^{(i)}$ and the weight vector $\bw$. Denote the solution as $\tilde\bA_1^{(i)},\ldots, \tilde\bA_q^{(i)}$}.
		\EndFor
	\end{algorithmic}
	\Return  $\tilde\bA_1^{(\texttt{iters})},\ldots,\tilde \bA_q^{(\texttt{iters})}$.
\end{algorithm}

%%%%%%%%%%%%%%%%%%%%%%%%%%%%%%%%%%%
%%%%%%%%%%%%%%%%%%%%%%%%%%%%%%%%%%%
%%%%%%%%%%%%%%%%%%%%%%%%%%%%%%%%%%%
\section{Numerical Experiments}\label{sec:numerics}
In this section, we illustrate RSMI on three numerical examples. 

\begin{table}
	\centering
	\begin{tabular}{|c|c|c|}\hline
		Example                                                             & $\lambda^{(0)}$ & \texttt{lr} \\ \hline
		\begin{tabular}[c]{@{}c@{}}Delay heat rod model \end{tabular}                                                        & $5\times 10^{-3}$         & $5\times 10^{-3}$                                \\ \hline
		Fishtail robot                                                      & $5\times 10^{-3}$         & $1\times 10^{-2}$                                        \\ \hline
		\begin{tabular}[c]{@{}c@{}}Parametric  thermal block\end{tabular} & $5\times 10^{-3}$         & $5\times 10^{-3}$                  \\         \hline                 
	\end{tabular}
	\caption{Parameters used in the numerical experiments.}
	\label{tab:parameters_examples}
\end{table}

\subsection{Implementation details} \label{sec:implementation_details}
We make use of the PyTorch library \cite{NEURIPS2019_9015} to compute gradients with automatic differentiation and optimize with \texttt{NAdam} \cite{dozat2016incorporating}. To compute $\|\cR_i\|$ in \eqref{eq:rRSMI}, we use a combination of $l_1$ and $l_2$-norms with an equal weight as in 
\begin{equation}
	\|\cR_i\| = \|\texttt{vect}(\cR_i)\|_{l_1} + \|\texttt{vect}(\cR_i)\|_{l_2}^2,
\end{equation}  
where $\texttt{vect}(\cdot)$ is the vectorization of a matrix. 

We compare models obtained with three different approaches: 
\paragraph{Benchmark approach} We construct a structured model with no regularization by setting $\lambda^{(0)} =0$ in \eqref{eq:rRSMI} and  
\texttt{iters} to $0$ in \Cref{algo:IterAlgo_WNN}. We take $50,000$ optimization iterations with an initial learning rate \texttt{lr} which is reduced by the factor of $5$ after every $12,500$-th update. This approach serves as our benchmark, and it is closely related to the approach discussed in \cite{schulze2018data} for the identification of structured systems. We denote this approach by \noreg~in the following.

\paragraph{RSMI with nuclear norm} We learn structured models with \eqref{eq:rRSMI} with the constant-weight nuclear norm and with regularization parameter as given in \Cref{tab:parameters_examples}. 
The rest of the setup is the same as in the previous paragraph. We denote this approach by \nnreg.
\paragraph{RSMI with weighted nuclear norm} In this approach, we employ a weighted nuclear norm in \eqref{eq:rRSMI}. We first construct a model in Step~1 in \Cref{algo:IterAlgo_WNN}  with $\lambda^{(0)} = \lambda_{\text{reg}}$ and learning rate given in \Cref{tab:parameters_examples}. 
Then, in Step 6 of \Cref{algo:IterAlgo_WNN} and the $i$-th iteration, we use $\lambda^{(i)} = \lambda^{(0)}/i$. A motivation to gradually decrease the regularization parameter is that we slowly give more weight to satisfying the interpolation conditions. But note that smaller singular values tend to stay small due to their high weight, which is inversely proportional to the singular value. Together with decreasing the regularization parameter, in Step 6, we also reduce the initial learning rate by a factor of two. Furthermore, \texttt{max\_val} in \Cref{algo:IterAlgo_WNN} is set to $10^4$. We denote this approach by \wnnreg.

\subsection{Delay heat rod model}
Consider the heating rod example with delay described in \cite{schulze2018data}. We construct a state-space model of order $n=101$ by setting the same parameters as in \cite{schulze2018data}. The system has delay $\tau = 1$. The transfer function of the system is of the form:
\begin{equation}\label{eq:delay_str}
	\bH(s) = \bC\left(s\bE-\bA -\bA_\tau e^{-\tau s}\right)^{-1}\bB. 
\end{equation}
For training data, we consider $\cN = 150$ points $\sigma_1, \dots, \sigma_{\cN}$ on the imaginary axis in the frequency range $[10^{-1}, 10^3]$ and the corresponding outputs $\bH(\sigma_1), \dots, \bH(\sigma_{\cN})$. Using these training data, we aim to learn a dynamical-system model with a transfer function that interpolates the training data. We construct interpolating models using the three approaches \noreg, \nnreg, and \wnnreg. 
\begin{figure}[!tb]
	\centering
	\begin{subfigure}[t]{0.46\textwidth}
		\includegraphics[width =\textwidth]{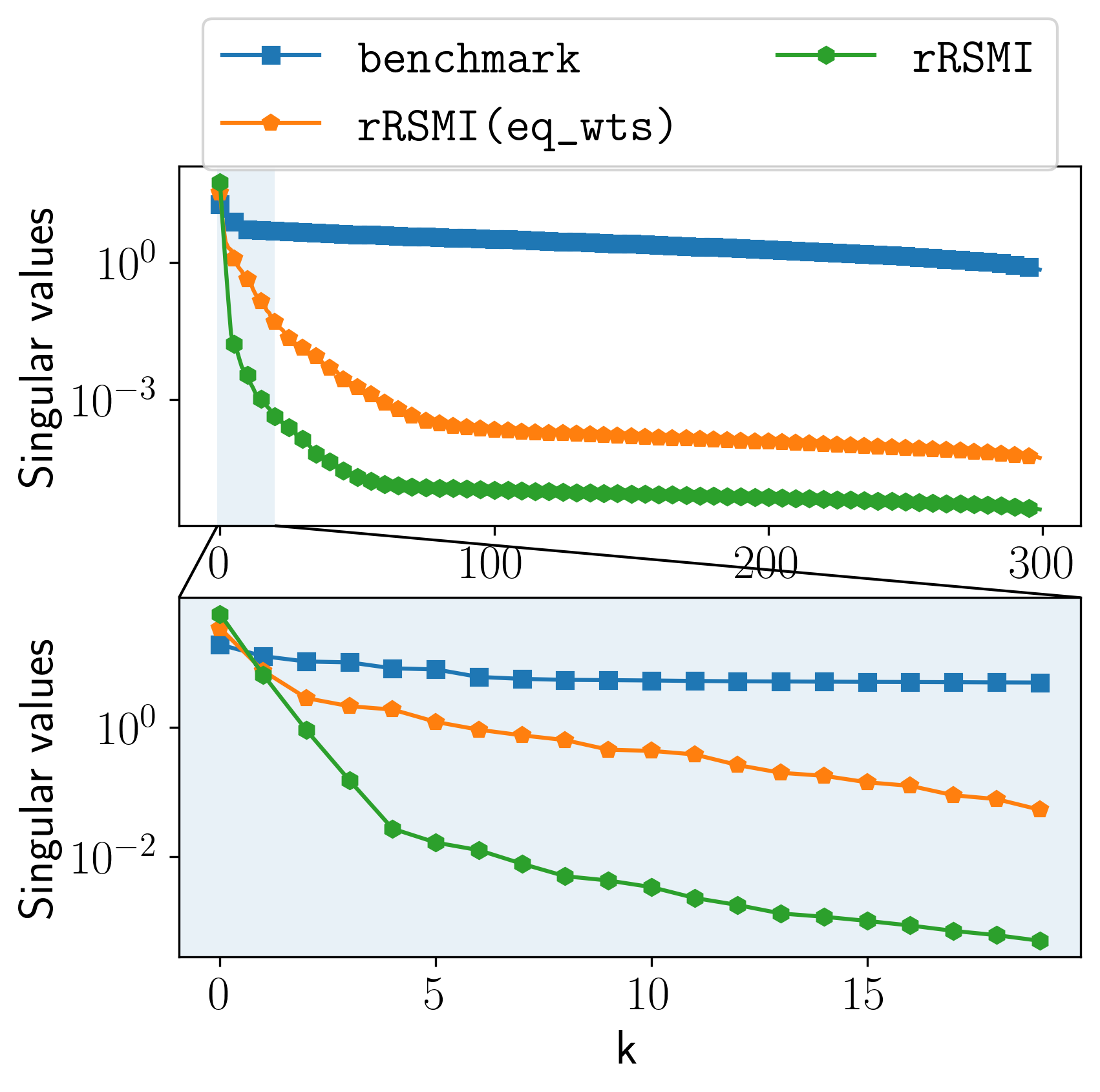}
		\caption{Decay of singular values.}
		\label{fig:Delay_decaySV}
	\end{subfigure}
	\begin{subfigure}[t]{0.53\textwidth}
		\includegraphics[width =\textwidth]{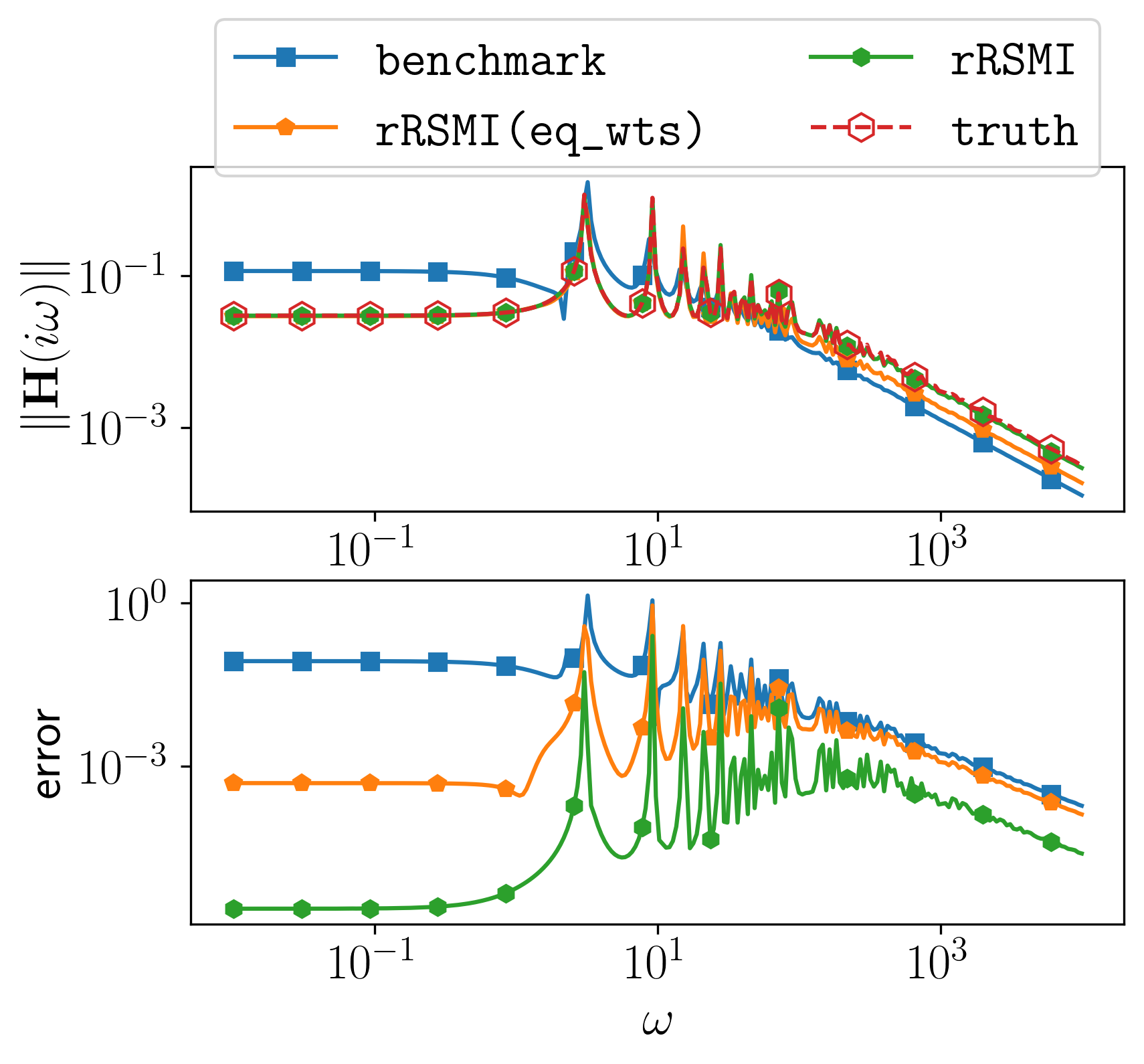}
		\caption{Transfer function comparison. }
		\label{fig:Delay_tfs}
	\end{subfigure}
	\caption{Delay heat rod problem: Plot (a) shows the decay of the singular values corresponding to the models obtained with \noreg, \nnreg, and \wnnreg. The singular values corresponding to the model obtained with \wnnreg~decays fastest, which means that it can be compressed most. Plot (b) shows the error of the learned models after compressing them to order $r = 3$, where the \wnnreg~model obtains the lowest error.} 
	\label{fig:Delay_svd_tf}
\end{figure}

\subsubsection{Results}\label{sec:THBResults}
In \cref{fig:Delay_decaySV}, we plot the singular values of the matrices of the models obtained by the three approaches. 
Note that all models aim to interpolate the training data and encode the same structure. The decay of the singular values corresponding to the models obtained from the three approaches indicates how much the models can be compressed, determining if there exist lower-order models that can also interpolate well. 

The singular values corresponding to the model obtained with \noreg~decay slowly, which implies that the model cannot be compressed well. 
In contrast, the model learned with our approach \nnreg~leads to a faster decay of the singular values, and thus the learned model can be compressed while approximating the training data well. 
The decay of the singular values is even faster for the model learned with \wnnreg~that relies on the weighted nuclear norm for regularization.  
This implies that the model obtained with \wnnreg~is the most compressible, resulting in the lowest order. 

We now construct low-order models by truncating the original models obtained with \noreg, \nnreg, and \wnnreg. We truncate after the first three most significant singular values. 
As a result, we obtain models of order three that have the same delay structure as \eqref{eq:delay_str}. We determine the quality of these models on test data, which are different from the training data. We consider $250$ frequency points in the interval $[10^{-2}, 10^4]$ on a logarithm scale, which is also outside of the domain of the training data. We compare the obtained truncated models with the ground truth in \Cref{fig:Delay_tfs}. The plots show that the model obtained with our approach \wnnreg~provides a low-dimensional model with the lowest error on the training and test data. \Cref{fig:delay_med_err} shows the median error over all test data points, where the model obtained with the proposed \wnnreg~approach achieves orders of magnitude lower errors than the model obtained with \noreg~approach that has no rank-minimization constraints.

\subsubsection{Imposing symmetry onto matrices} 
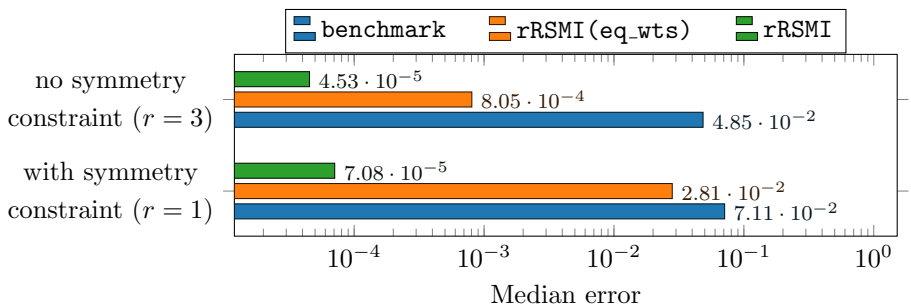
\begin{figure}
	\begin{tikzpicture}
		\begin{axis}[ 
			height=4cm,
			width=5cm,
			% ybar=0.5pt, % change this of you have more than one bar per column
			x=0.75cm, % just play with the relation of these
			bar width=0.2cm, % just play with with the relation of these
			enlarge y limits={abs=0.60cm},
			enlarge x limits={abs=1cm},
			% bar shift=2,
			xbar, xmin=0, xmax = 0.4,
			xlabel={Median error},
			xmode=log,
			symbolic y coords={%
				{approch1},
				{approch2},
			},
			ytick=data,
			yticklabels={\begin{tabular}[c]{@{}c@{}}no symmetry \\ constraint $(r=3)$\end{tabular},
				\begin{tabular}[c]{@{}c@{}}with symmetry \\ constraint $(r=1)$\end{tabular},
			},
			y dir=reverse,
			nodes near coords={\pgfmathprintnumber{\pgfplotspointmeta}},
			nodes near coords align={horizontal},
			nodes near coords style={font=\footnotesize},
			log origin=infty,
			point meta=rawx,
			legend style={at={(0.5,1.25)}, anchor=north, legend columns=-1, /tikz/every even column/.append style={column sep=0.5cm}},
			]
			
			\addplot[matplotlibcolor1!20!black,fill=matplotlibcolor1] coordinates {
				(4.85e-2,{approch1}) 
				(7.11e-2,{approch2})
			};
			
			\addplot[matplotlibcolor2!20!black,fill=matplotlibcolor2] coordinates {
				(8.05e-4,{approch1}) 
				(2.81e-2,{approch2}) 
			};
			
			\addplot[matplotlibcolor3!10!black,fill=matplotlibcolor3] coordinates {
				(4.53e-5,{approch1}) 
				(7.08e-5,{approch2}) 
			};
			
			\legend{\noreg, \nnreg, \wnnreg}
		\end{axis}
	\end{tikzpicture} %
	\caption{Delay heat rod model: The plot shows that models learned with our approach \wnnreg~achieve orders of magnitude lower median errors over the test data points.}
	\label{fig:delay_med_err}
\end{figure}

\begin{figure}[!tb]
	\centering
	\begin{subfigure}[t]{0.46\textwidth}
		\includegraphics[width =\textwidth]{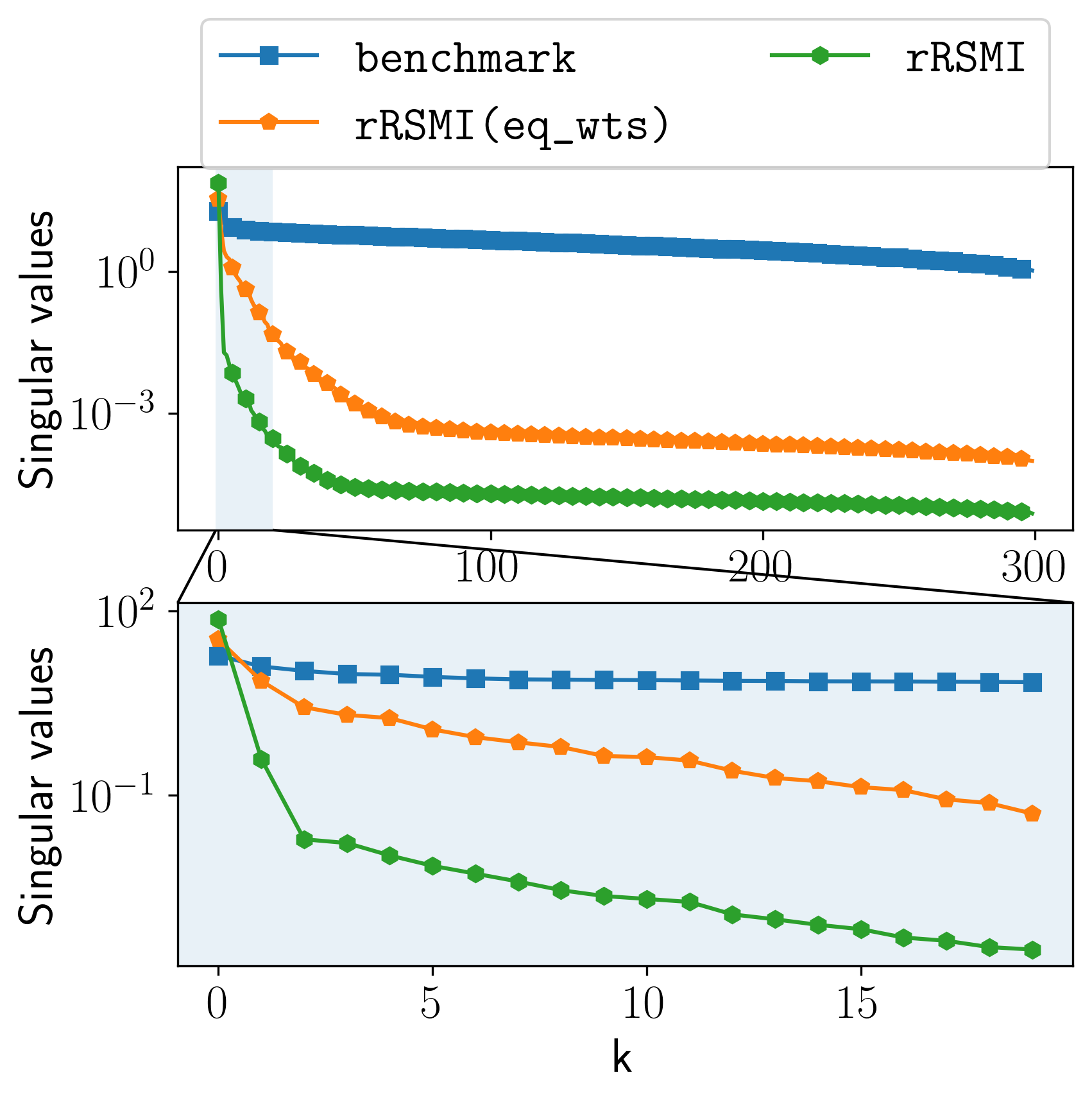}
		\caption{Decay of singular values.}
		\label{fig:Delay_Symm_decaySV}
	\end{subfigure}
	\begin{subfigure}[t]{0.53\textwidth}
		\includegraphics[width =\textwidth]{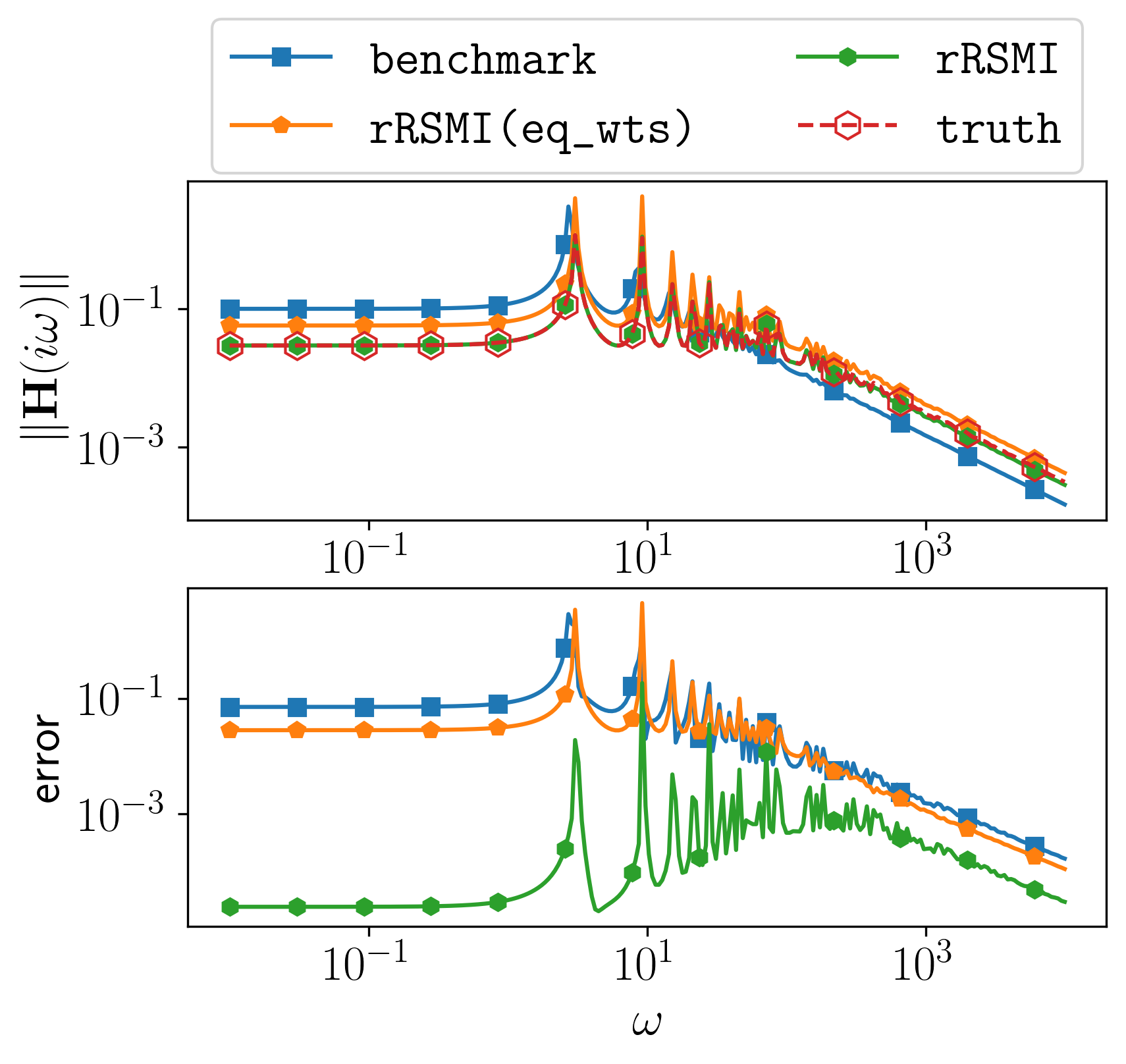}
		\caption{Transfer function comparison.}
		\label{fig:Delay_Symm_tfs}
	\end{subfigure}
	\caption{Delay heat rod model: Plot (a) shows that imposing symmetry leads to a faster decay of the singular values corresponding to the model learned with our approach \wnnreg, compared to not imposing symmetry as shown in \Cref{fig:Delay_svd_tf}. The faster decay means that the model can be compressed to just one dimension $r = 1$ and still provides accurate predictions on test data, as shown in plot~(b).} 
	\label{fig:Delay_Symm_svd_tf}
\end{figure}

We repeat the analog experiments to \Cref{sec:THBResults} but impose symmetry as described in \Cref{subsec:imposesymm}. 
We first examine the decay of singular values shown in \Cref{fig:Delay_Symm_decaySV}, which again indicates a similar trend as in the non-symmetric case. 
The first singular vector corresponding to the model obtained with \wnnreg~captures more than $99\%$ energy. Thus, we construct one-dimensional models by truncating the models obtained with \noreg, \nnreg, and \wnnreg~after the first mode and compare their performance on the test data in \Cref{fig:Delay_Symm_tfs}. Our approach \wnnreg~yields the model with the lowest median error, which is below $10^{-4}$ and shown in \Cref{fig:delay_med_err}.

\begin{figure}[!tb]
	\centering
	\begin{subfigure}[t]{0.5\textwidth}
		\includegraphics[width =\textwidth]{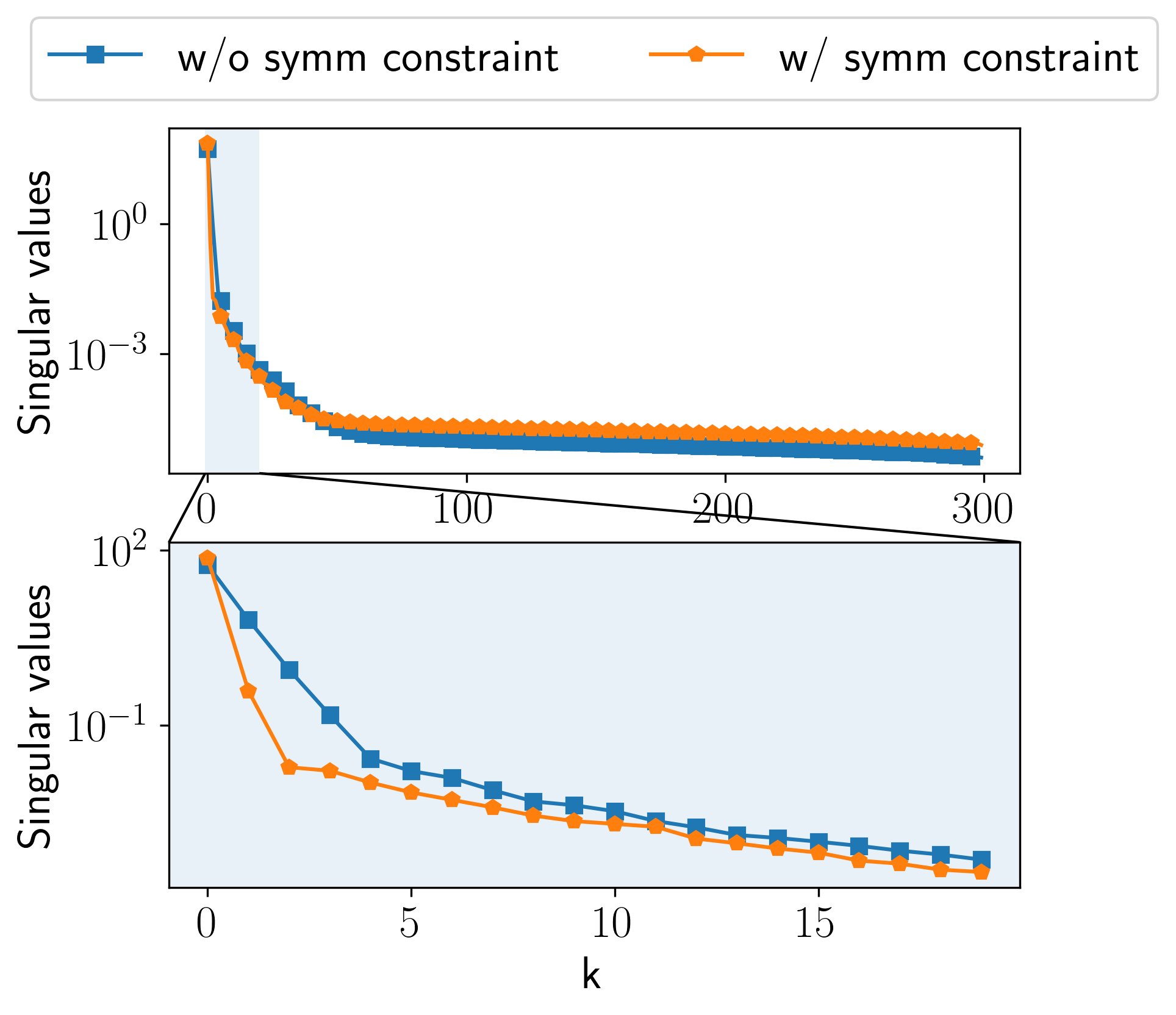}
		\caption{Decay of singular values.}
		\label{fig:Delay_SymmvsNonsymm_decaySV}
	\end{subfigure}
	\begin{subfigure}[t]{0.49\textwidth}
		\includegraphics[width =\textwidth]{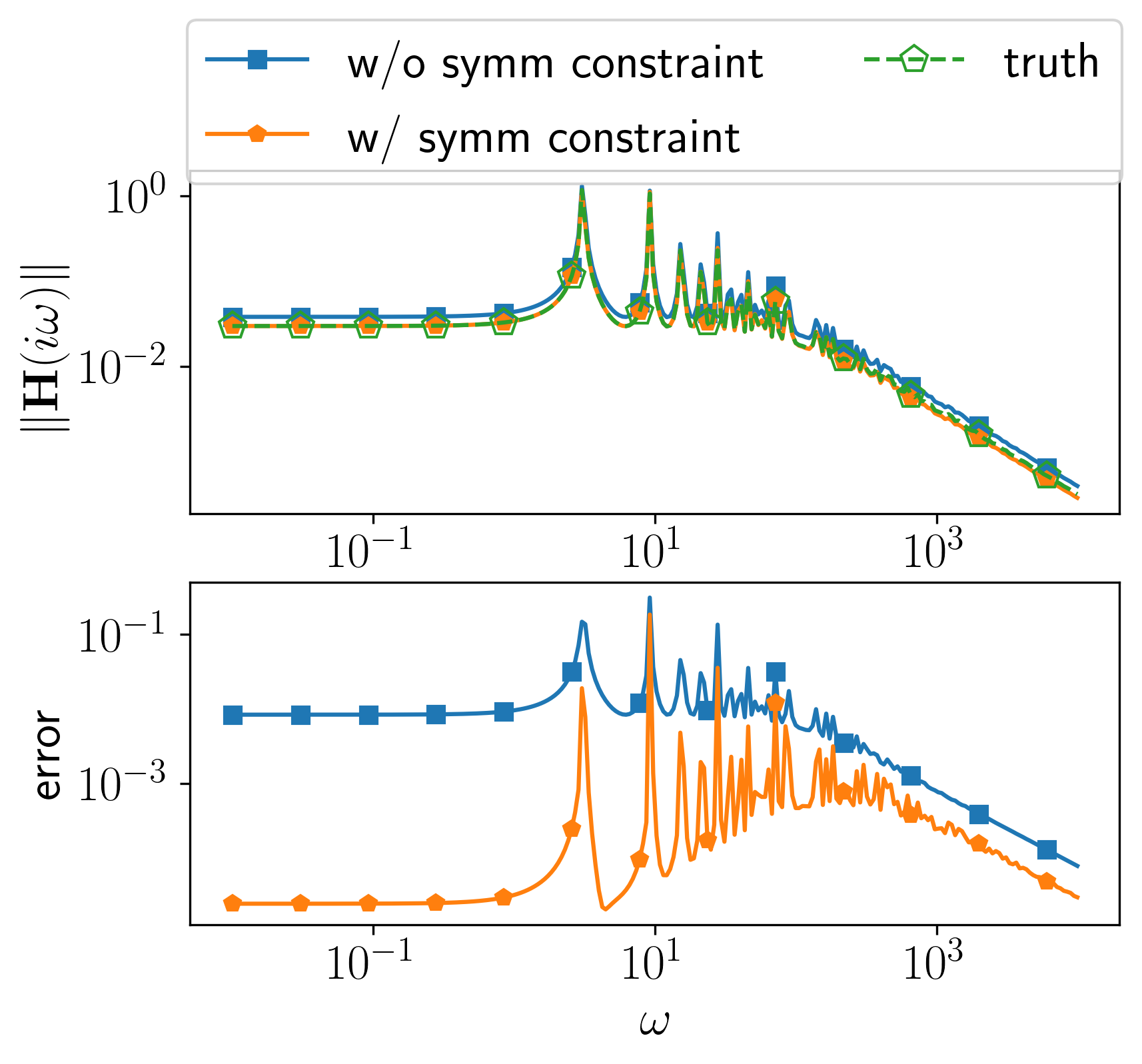}
		\caption{Transfer function comparison.}
		\label{fig:Delay_SymmvsNonsymm_comparison_tfs}
	\end{subfigure}
	\caption{Delay heat rod model: The plots show the faster decay of the singular values corresponding to models learned with \wnnreg~with symmetry constraints. Compression to just one dimension leads to a model with two orders of magnitude lower error on test data if symmetries are enforced.} 
	\label{fig:Delay_SymmvsNonsymm_svd_tf}
\end{figure}

We explicitly compare the performance of the models that impose symmetry versus the models obtained without symmetry constraints. We do this only for the models obtained with the \wnnreg~approach. 
What we observe is a faster decay of singular values of the models obtained with \wnnreg~when the symmetry in matrices is imposed, see \Cref{fig:Delay_SymmvsNonsymm_decaySV}. Therefore, we obtain lower dimensional models with \wnnreg~in this example with symmetry constraints. We now compare the model obtained with truncating after the first mode with symmetry imposed to the truncated model without symmetry; see 
\Cref{fig:Delay_SymmvsNonsymm_comparison_tfs}. We observe that the symmetry constraint leads to almost two orders of magnitude lower error in this example. 

\subsection{Fishtail robot}
\begin{figure}[!tb]
	\centering
	\begin{subfigure}[b]{0.45\textwidth}
		\includegraphics[width = \textwidth]{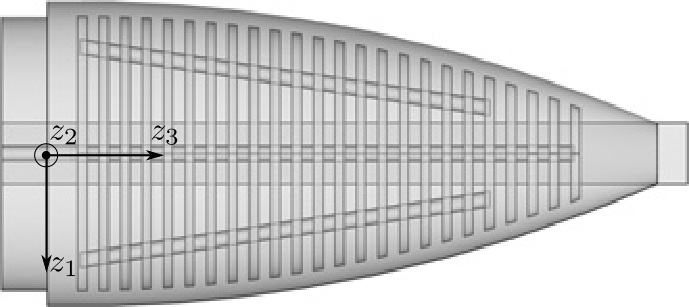}
		\caption{A semantic figure of the fishtail robot from \cite{morSaaSW19}.}
		\label{fig:fishtail_semantic_1}
	\end{subfigure}
	\begin{subfigure}[b]{0.45\textwidth}
		\includegraphics[width = \textwidth]{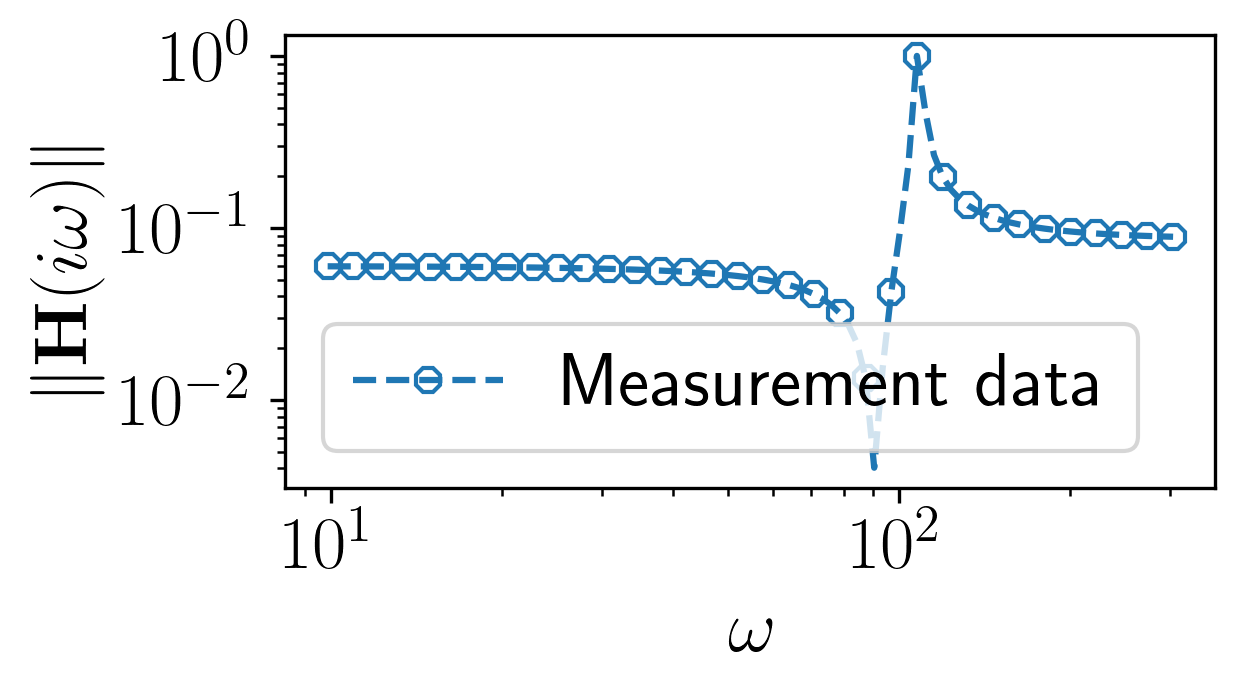}
		\caption{Training data.}
		\label{fig:fishtail_semantic_2}
	\end{subfigure}
	\caption{Fishtail robot: The plots show a sketch of the robot and the training data.}
	\label{fig:fishtail_semantic}
\end{figure}

In our second example, we consider an artificial fishtail robotic model from \cite{morSaaSW19} as shown in \Cref{fig:fishtail_semantic_1}. The state-space model from which we generate data has a second-order structure of the form:
\begin{equation}
	\begin{aligned}
		\bM\ddot{\bx}(t) + \bD \dot{\bx}(t) + \bK \bx(t) &= \bB\bu(t),\\
		\by(t) &= \bC\bx(t),
	\end{aligned}
\end{equation}
and its state has dimension $\approx 700,000$. In our setting, we have only available  $100$ measurements of the transfer function at the frequencies in the interval $[10^{1}, 3\times 10^{2}]$ as shown in \Cref{fig:fishtail_semantic_2}. We have normalized the transfer function by dividing by the maximum absolute value; otherwise $\|\bH(s)\|$ is in the range of $ 10^{-7}$ and thus becomes numerically too small for the considered frequency range.

\subsubsection{Results} We leverage that transfer functions of second-order systems have the form
\begin{equation}
	\bH_{\texttt{so}} = \bC\left(s^2\bM+ s\bD + \bK\right)^{-1}\bB.
\end{equation}
We employ the \noreg, \nnreg, and \wnnreg~approaches to construct interpolating models. Before doing so, we scale the function $s^2$ and $s$ by factor $\gamma_{\bM}$ and $\gamma_\bD$. The reason for this is that while assembling the matrices for $s^2$ when the frequency interval is $[10^{1}, 3\times 10^{2}]$, the corresponding matrix becomes dominating, especially for high frequencies. As a result, the optimization problems become computationally harder to solve.  
In such scenarios, scaling can help the optimizer. However, note that the scaling does not affect the transfer function, as it is invariant under scaling, 
\begin{equation}
	\bH_{\texttt{so}} = \bC\left(s^2\bM+ s\bD + \bK\right)^{-1}\bB = \bC\left(\gamma_{\bM} s^2\tilde \bM+ \gamma_{\bD} s\tilde \bD + \tilde \bK\right)^{-1}\bB,
\end{equation}
where $\tilde \bM = \bM/\gamma_{\bM}$, $\tilde\bD = \bD/\gamma_\bD$ and $\tilde\bK = \bK$. Therefore, we seek to learn $\tilde\bM,\tilde\bD$ and $\tilde\bK$ with the scaled $\alpha_1(s) = \gamma_{\bM}s^2$ and $\alpha_2(s) = \gamma_{\bD}s$. For this example, we choose $\gamma_{\bM} = 10^{-3}$ and $\gamma_\bD = 10^{-1.5}$.  We identify the interpolating realizations using all three approaches, as in the previous example. 

\begin{figure}[!tb]
	\centering
	\begin{subfigure}[t]{0.46\textwidth}
		\includegraphics[width =1\textwidth]{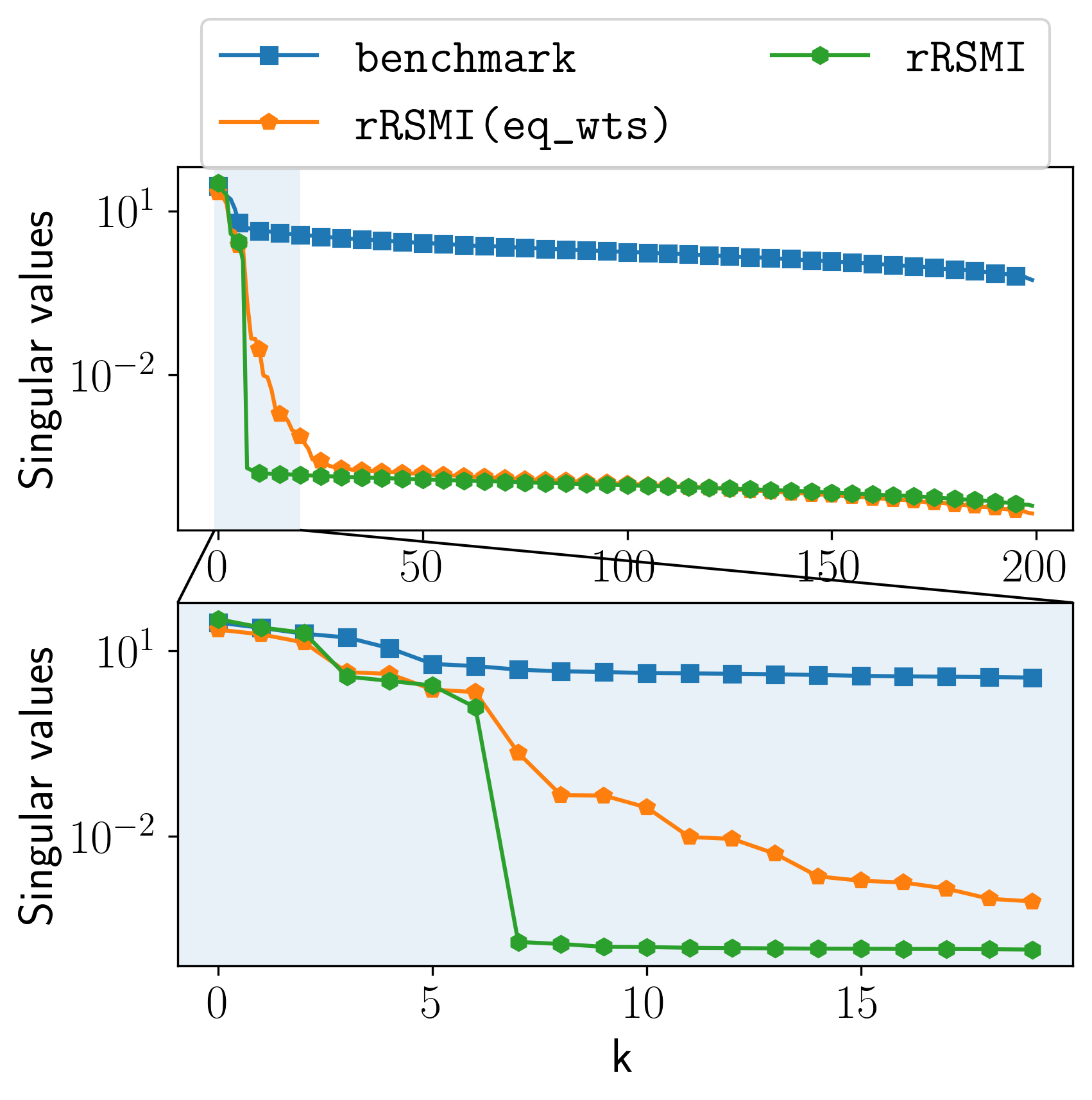}
		\caption{The decay of singular values.}
		\label{fig:fish_decaySV}
	\end{subfigure}
	\begin{subfigure}[t]{0.53\textwidth}
		\includegraphics[width = 1\textwidth]{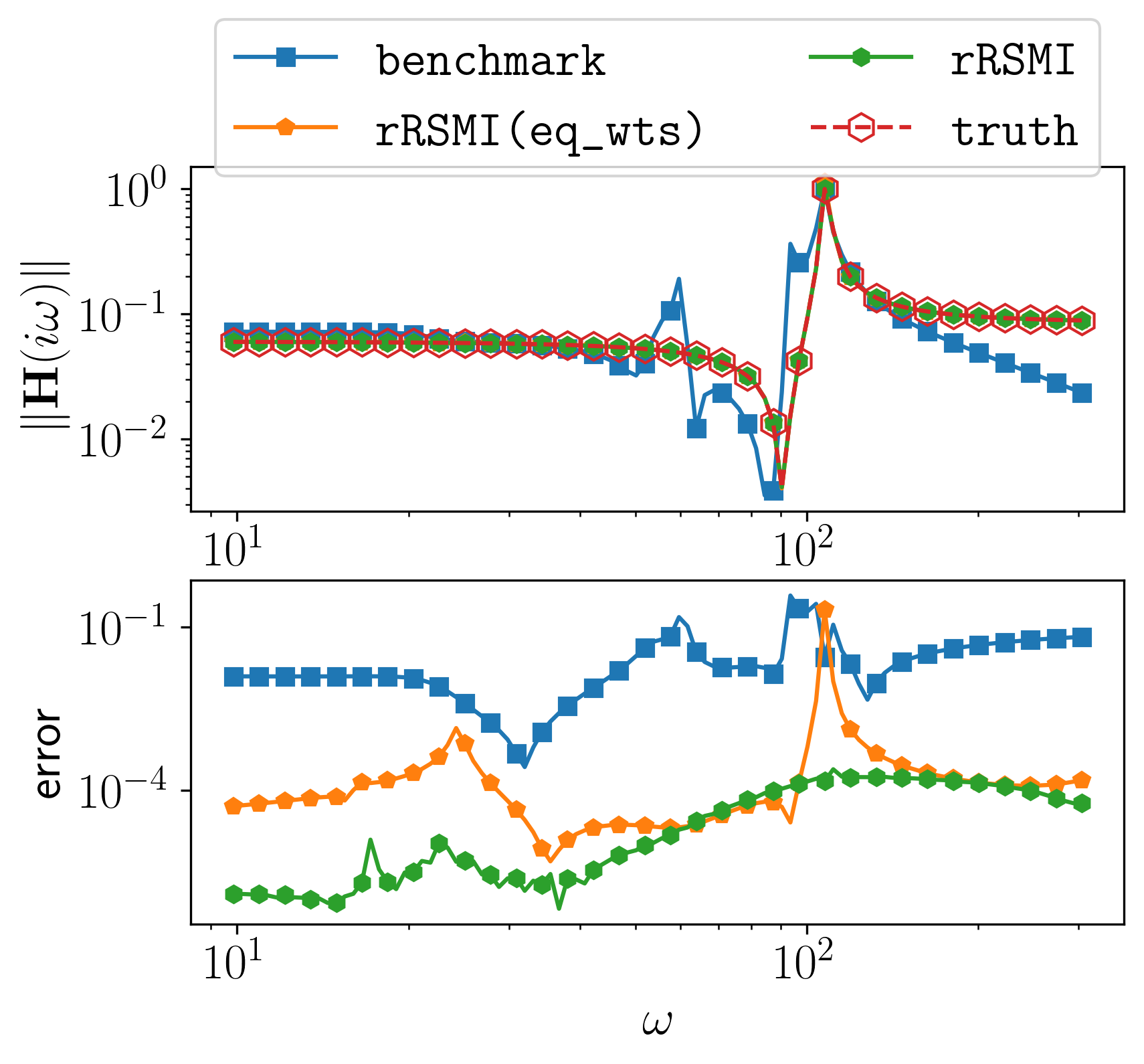}
		\caption{Transfer function comparison.}
		\label{fig:fish_tfs}
	\end{subfigure}
	\caption{Fishtail Robot: Plot (a) shows that the singular values corresponding to the model learned with our approach \wnnreg~decay fastest, which indicates that the model can be compressed well. Plot (b) shows the prediction error on data of the learned models when compressed to order $r = 7$, where our approach leads to the lowest error.}
	\label{fig:fishtail_svd_tf}
\end{figure}

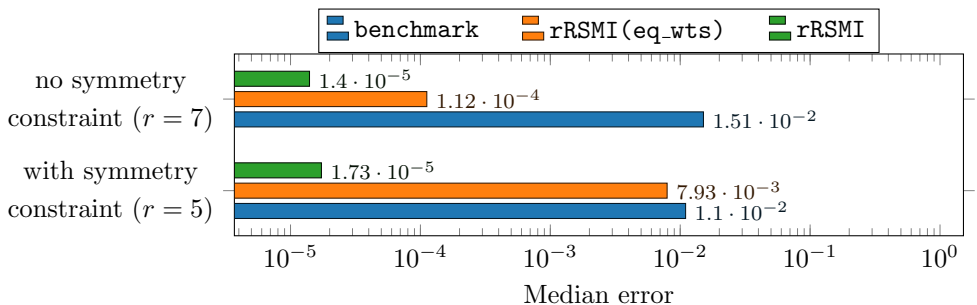
\begin{figure}
	\begin{tikzpicture}
		\begin{axis}[ 
			height=4cm,
			width=5cm,
			% ybar=0.5pt, % change this of you have more than one bar per column
			x=0.75cm, % just play with the relation of these
			bar width=0.2cm, % just play with with the relation of these
			enlarge y limits={abs=0.60cm},
			enlarge x limits={abs=1cm},
			% bar shift=2,
			xbar, xmin=0, xmax = 0.4,
			xlabel={Median error},
			xmode=log,
			symbolic y coords={%
				{approch1},
				{approch2},
			},
			ytick=data,
			yticklabels={\begin{tabular}[c]{@{}c@{}}no symmetry \\ constraint $(r=7)$\end{tabular},
				\begin{tabular}[c]{@{}c@{}}with symmetry \\ constraint $(r=5)$\end{tabular},
			},
			y dir=reverse,
			nodes near coords={\pgfmathprintnumber{\pgfplotspointmeta}},
			nodes near coords align={horizontal},
			nodes near coords style={font=\footnotesize},
			log origin=infty,
			point meta=rawx,
			legend style={at={(0.5,1.25)}, anchor=north, legend columns=-1, /tikz/every even column/.append style={column sep=0.5cm}},
			]
			
			\addplot[matplotlibcolor1!20!black,fill=matplotlibcolor1] coordinates {
				(1.51e-2,{approch1})
				(1.10e-2,{approch2}) 
			};
			
			\addplot[matplotlibcolor2!20!black,fill=matplotlibcolor2] coordinates {
				(1.12e-4,{approch1}) 
				(7.93e-3,{approch2}) 
			};
			
			\addplot[matplotlibcolor3!10!black,fill=matplotlibcolor3] coordinates {
				(1.40e-5,{approch1}) 
				(1.73e-5,{approch2}) 
			};
			
			\legend{\noreg, \nnreg, \wnnreg}
		\end{axis}
	\end{tikzpicture} %
	\caption{Fishtail robot: The median error on the data is orders of magnitude lower for models learned with our approach \wnnreg~than without rank-minimization as in the \noreg~approach.} 
	\label{fig:fishrobot_med_err}
\end{figure}

We plot the singular values corresponding to the learned models in \Cref{fig:fish_decaySV}. The singular values exhibit similar behavior as in the previous example, in the sense that the singular values corresponding to the model learned with our \wnnreg~approach decay fastest.   
This indicates that the model learned with \wnnreg~can be compressed efficiently to a lower-order model without significant loss of accuracy. 
We truncate after the first $r=7$ dominating singular values and so obtain models of order seven. We compare the truncated models on the training data in \Cref{fig:fish_tfs,fig:fishrobot_med_err}. Again, \wnnreg~yields the model with the lowest median error among all models, showing the robustness of the \wnnreg~approach to learn lower-order models that approximately interpolate the data. 

\subsubsection{Imposing symmetry}
\begin{figure}[!tb]
	\centering
	\begin{subfigure}[t]{0.46\textwidth}
		\includegraphics[width =1\textwidth]{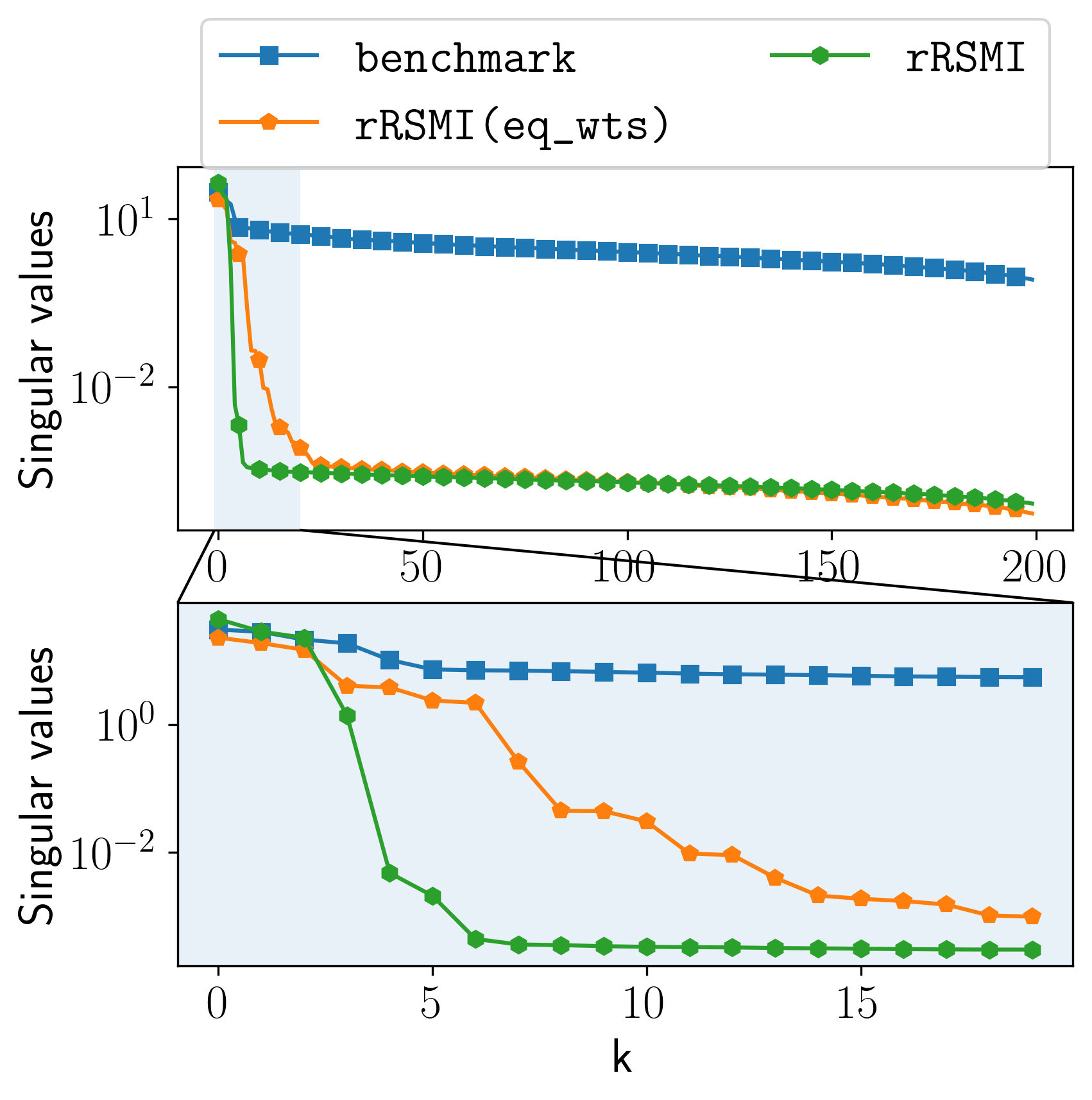}
		\caption{Decay of singular values.}
		\label{fig:fish_symm_decaySV}
	\end{subfigure}
	\begin{subfigure}[t]{0.53\textwidth}
		\includegraphics[width = 1\textwidth]{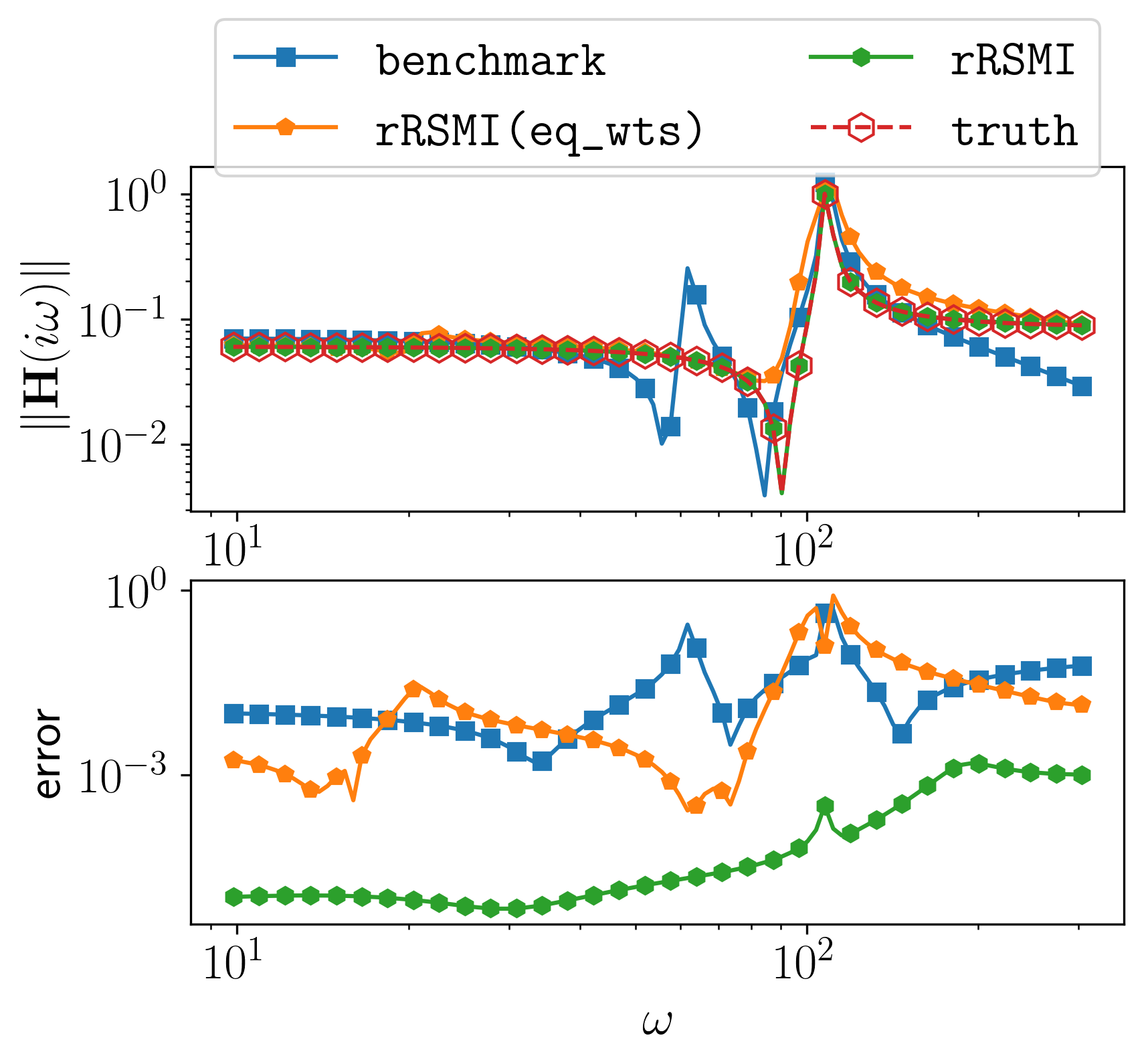}
		\caption{Transfer function comparison.}
		\label{fig:fish_symm_tfs}
	\end{subfigure}
	\caption{Fishtail robot: Plot (a) shows that imposing symmetry helps to achieve a faster decay of the singular values corresponding to the learned models, which means the models can be compressed while remaining predictive. Plot (b) shows after compression to order $r = 5$, the model obtained with our approach \wnnreg~achieves the lowest error in this experiment.}
	\label{fig:fishtail_symm_svd_tf}
\end{figure}

Having symmetry in the matrices $\bM,\bD$, and $\bK$ is of high interest in mechanical systems. We, therefore, enforce symmetry in these matrices while learning them with the \noreg, \nnreg, and \wnnreg~approaches. The rest of the setup is the same as in the previous experiment. The singular values corresponding to the models learned with symmetry constraints are shown in \Cref{fig:fish_symm_decaySV}. Again, the model obtained with \wnnreg~shows the fastest decay of the singular values. Moreover, with symmetry imposed, the decay is even faster than in the case without symmetry. Consequently, we truncate earlier and obtain a model of lower order than when symmetry is not imposed. We construct models of the order $r=5$ using all three methods \noreg, \nnreg, \wnnreg~by projection using the corresponding dominant subspaces spanned by the left singular vectors and compare the error of the truncated models in \Cref{fig:fishrobot_med_err,fig:fish_symm_tfs}. Our proposed approach \wnnreg~leads to the model with the lowest median error by more than one order of magnitude. 

\subsection{Thermal block example}
We now consider a $2\times 2$ thermal block, whose solution is described by the Poisson equation with the following diffusion coefficient:
\begin{equation}
	\begin{aligned}
		\eta(\zeta,p) = p_1\cX_{\left(0,\tfrac{1}{2}\right)^2}(\zeta) + p_2\cX_{\left(0,\tfrac{1}{2}\right)\times \left(\tfrac{1}{2},1\right)}(\zeta) + p_3\cX_{\left(\tfrac{1}{2},1\right)^2}(\zeta) + p_4\cX_{\left(0,\tfrac{1}{2}\right)\times\left(\tfrac{1}{2},1\right)}(\zeta),
	\end{aligned}
\end{equation}
where $p_i \in [0.1,10], i\in \{1,\ldots,4\}$ and $\cX_X$ is the indicator function of the set $X$. This is a widely used benchmark example \cite{morRavS21} and has also been studied in \cite{morMliG22}.  Having discretized the Poisson equation, we get a model of the form:
\begin{equation}
	\begin{aligned}
		\left(\bA_0 + p_1\bA_1 + p_2\bA_2 + p_3\bA_3 + p_4\bA_4\right)\bx(p) &= \bB,\\
		\by(p) = \bC\bx(p).
	\end{aligned}
\end{equation}
For more details on the governing equation, we refer to \cite{morMliG22,morRavS21}. We collect the data by considering four equidistant points in $\left[0.1, 10\right]$ for each $p_i, i = 1, \dots, 4$. 
Consequently, we get a total of $4^4 = 256$ training data points. For the construction of the models, we uniformly draw $200$ data points from the training data and determine a suitable order of the lower-order models by validating the learned models on the $56$ left-out data points. We additionally construct another test data set for this example by considering five equidistant points in $\left[0.1, 10\right]$ for each $p_i, i = 1, \dots 4$, which leads to a total of $5^4 = 625$ test points.

\subsubsection{Results} 
\begin{figure}[!tb]
	\centering
	\begin{subfigure}[t]{0.46\textwidth}
		\includegraphics[width =1\textwidth]{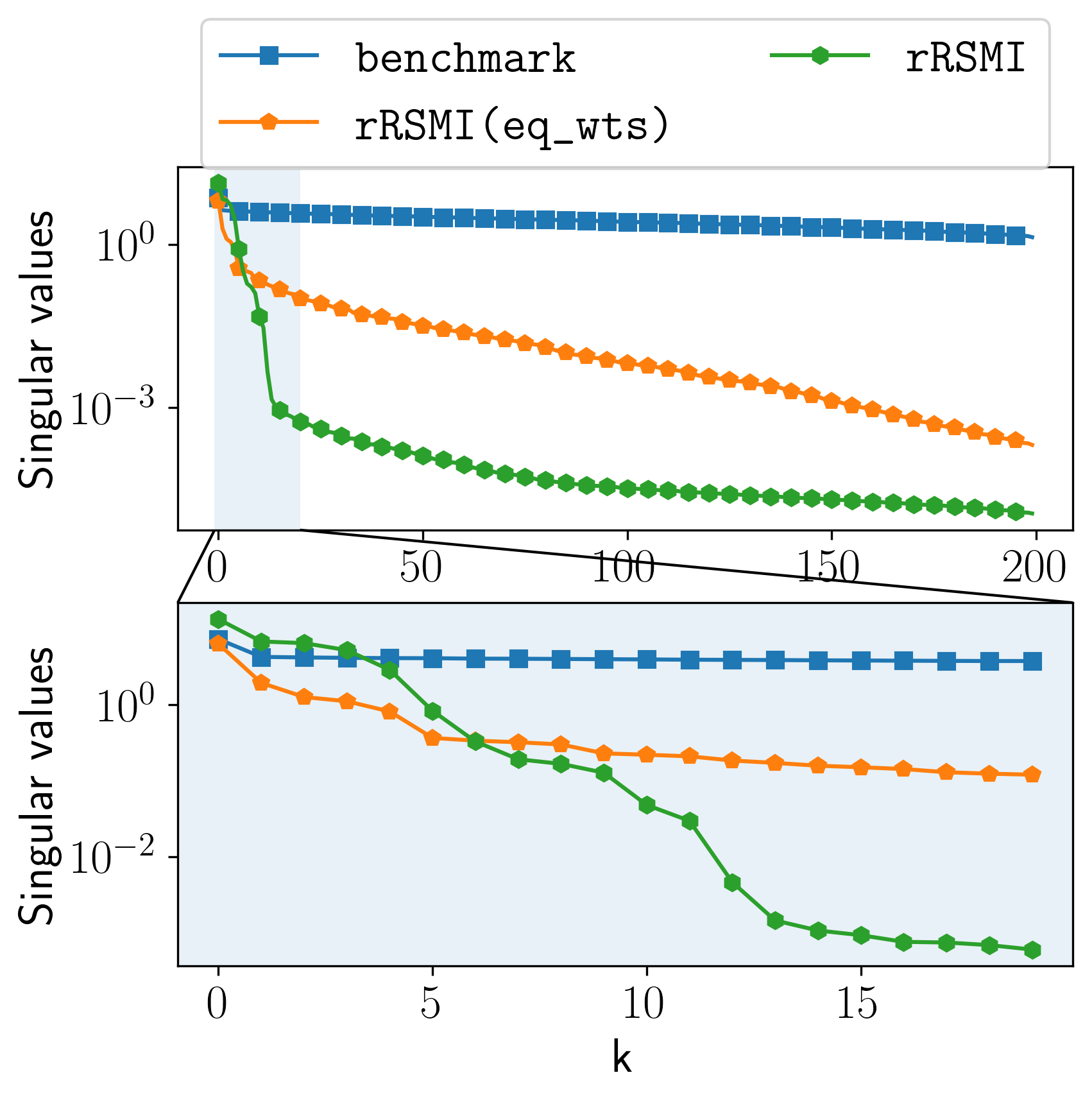}
		\caption{The decay of singular values.}
		\label{fig:thermal_decaySV}
	\end{subfigure}
	\begin{subfigure}[t]{0.53\textwidth}
		\includegraphics[width = 1\textwidth]{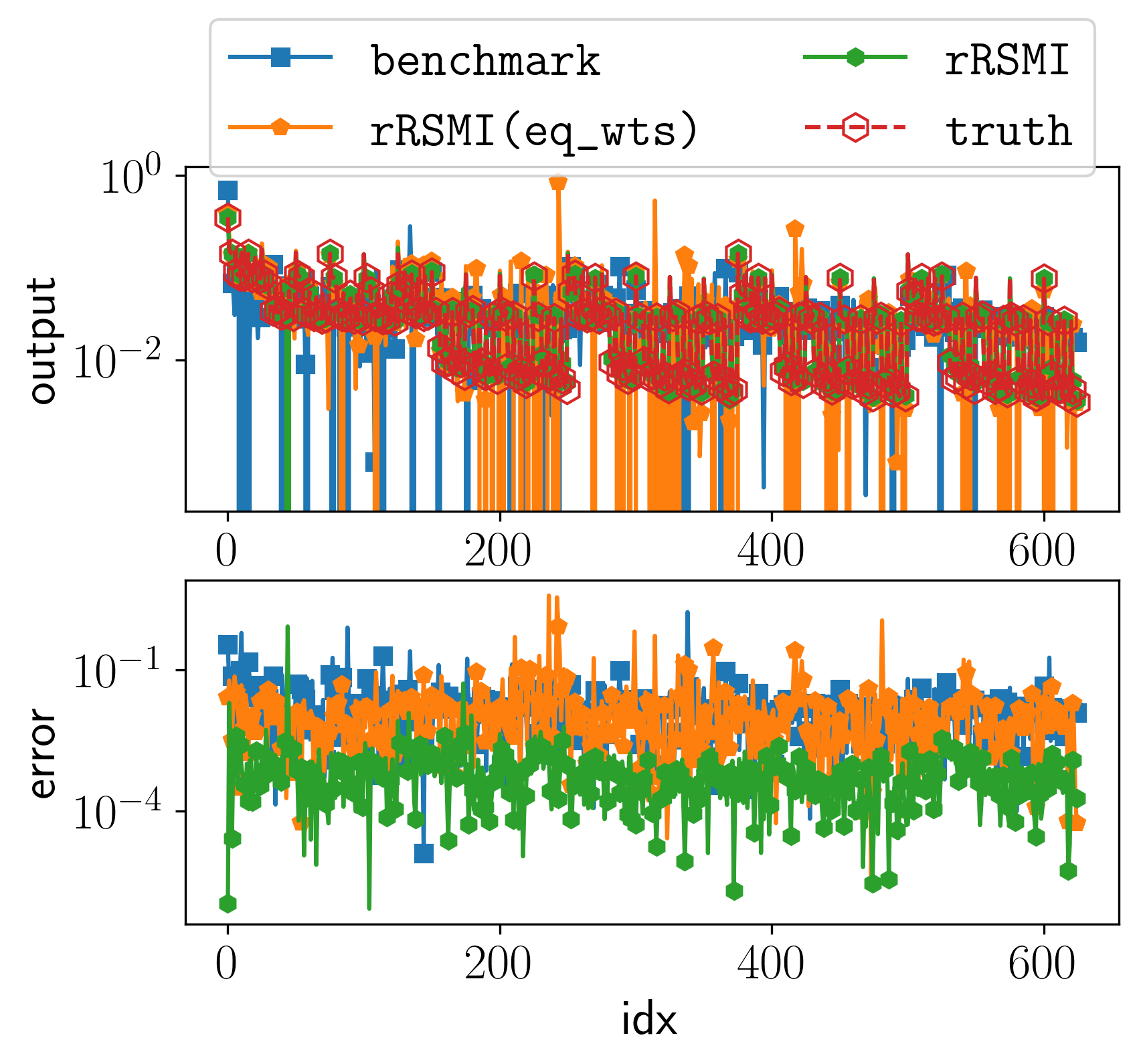}
		\caption{Output comparison on the testing data.}
		\label{fig:thermal_tfs}
	\end{subfigure}
	\caption{Thermal block example: Plot (a) shows the decay of the singular values, which is fastest for the model learned with \wnnreg. Plot (b) shows the error on test data after compression to order $r = 10$, where the \wnnreg~model achieves the lowest error.} 
	\label{fig:thermal_svd_tf}
\end{figure}

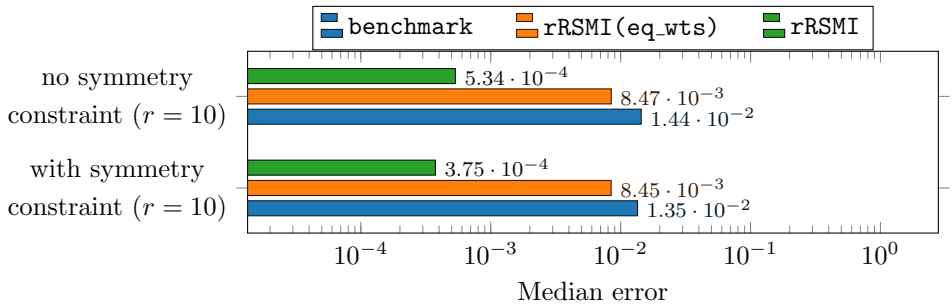
\begin{figure}
	\begin{tikzpicture}
		\begin{axis}[ 
			height=4cm,
			width=8cm,
			% ybar=0.5pt, % change this of you have more than one bar per column
			x=0.75cm, % just play with the relation of these
			bar width=0.2cm, % just play with with the relation of these
			enlarge y limits={abs=0.60cm},
			enlarge x limits={abs=2.5cm},
			% bar shift=2,
			xbar, xmin=0, xmax = 0.1,
			xlabel={Median error},
			xmode=log,
			symbolic y coords={%
				{approch1},
				{approch2},
			},
			ytick=data,
			yticklabels={\begin{tabular}[c]{@{}c@{}}no symmetry \\ constraint $(r=10)$\end{tabular},
				\begin{tabular}[c]{@{}c@{}}with symmetry \\ constraint $(r=10)$\end{tabular},
			},
			y dir=reverse,
			nodes near coords={\pgfmathprintnumber{\pgfplotspointmeta}},
			nodes near coords align={horizontal},
			nodes near coords style={font=\footnotesize},
			log origin=infty,
			point meta=rawx,
			legend style={at={(0.5,1.25)}, anchor=north, legend columns=-1, /tikz/every even column/.append style={column sep=0.5cm}},
			]
			
			\addplot[matplotlibcolor1!20!black,fill=matplotlibcolor1] coordinates {
				(1.44e-2,{approch1})
				(1.35e-2,{approch2}) 
			};
			
			\addplot[matplotlibcolor2!20!black,fill=matplotlibcolor2] coordinates {
				(8.47e-3,{approch1}) 
				(8.45e-3,{approch2}) 
			};
			
			\addplot[matplotlibcolor3!10!black,fill=matplotlibcolor3] coordinates {
				(5.34e-4,{approch1}) 
				(3.75e-4,{approch2}) 
			};
			
			\legend{\noreg, \nnreg, \wnnreg}
		\end{axis}
	\end{tikzpicture} %
	\caption{Thermal block example: The proposed approach \wnnreg~leads to models that achieve almost two orders of magnitude lower median errors on test data compared to models obtained with rank minimization as in \noreg.} 
	\label{fig:thermal_med_err}
	
\end{figure}

We employ the \noreg, \nnreg, and \wnnreg~approaches to construct models using the training data. The singular values are shown in \Cref{fig:thermal_decaySV}, which exhibit similar decay behavior as in the previous example: the model obtained with \wnnreg~achieves a faster decay than the models obtained with either \noreg~or \nnreg. We compress the model to order $r = 10$ and show the error of the models in \Cref{fig:thermal_tfs}. The median errors on the test data for realizations obtained using \noreg, \nnreg, and \wnnreg~are reported in \Cref{fig:thermal_med_err}. 
The \wnnreg~approach again yields a model that achieves an order of magnitude lower median error than the other approaches \nnreg~and \noreg.  

\subsubsection{Imposing symmetry}
\begin{figure}[!tb]
	\centering
	\begin{subfigure}[t]{0.46\textwidth}
		\includegraphics[width =1\textwidth]{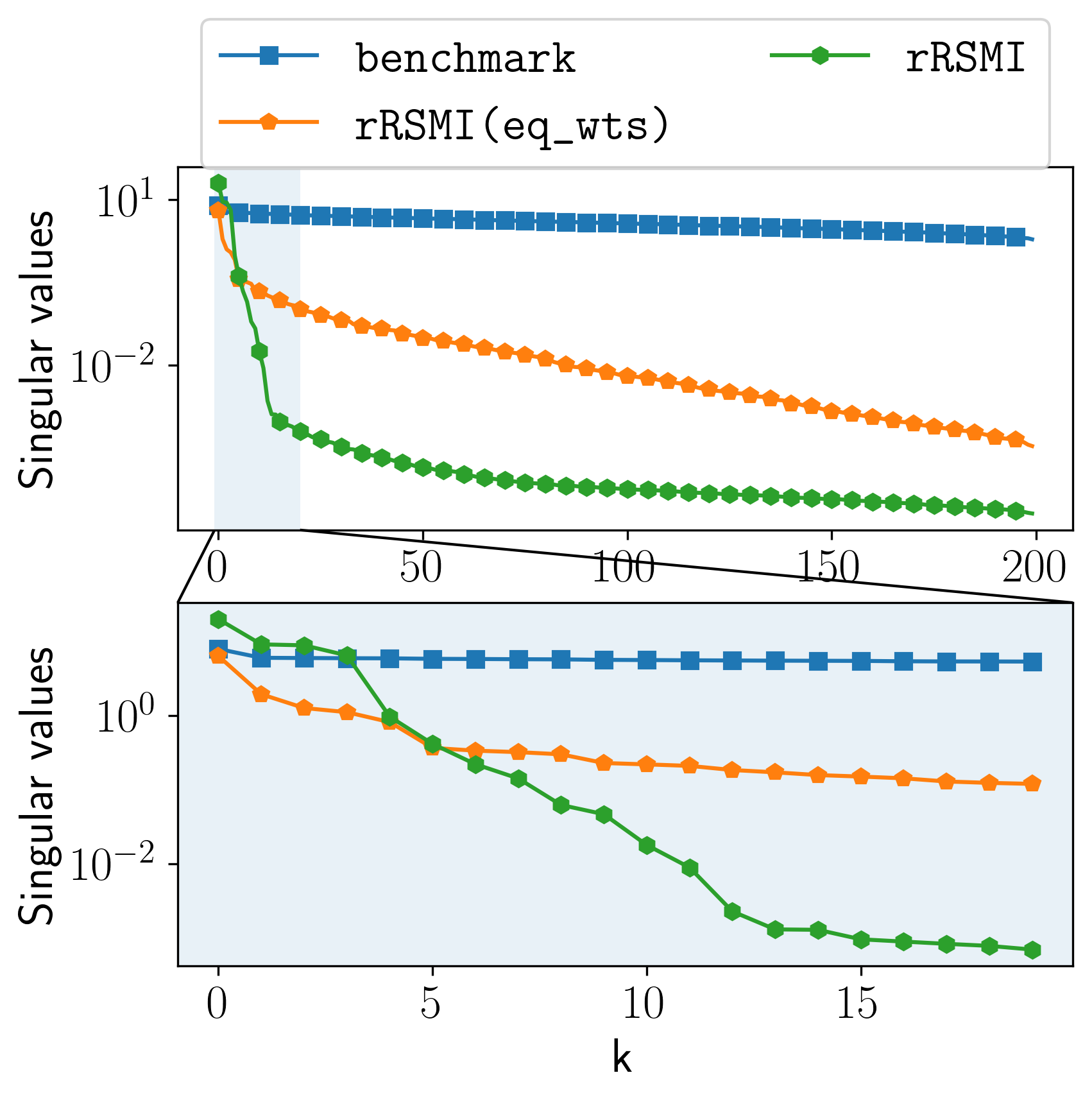}
		\caption{Decay of singular values.}
		\label{fig:thermal_symm_decaySV}
	\end{subfigure}
	\begin{subfigure}[t]{0.53\textwidth}
		\includegraphics[width = 1\textwidth]{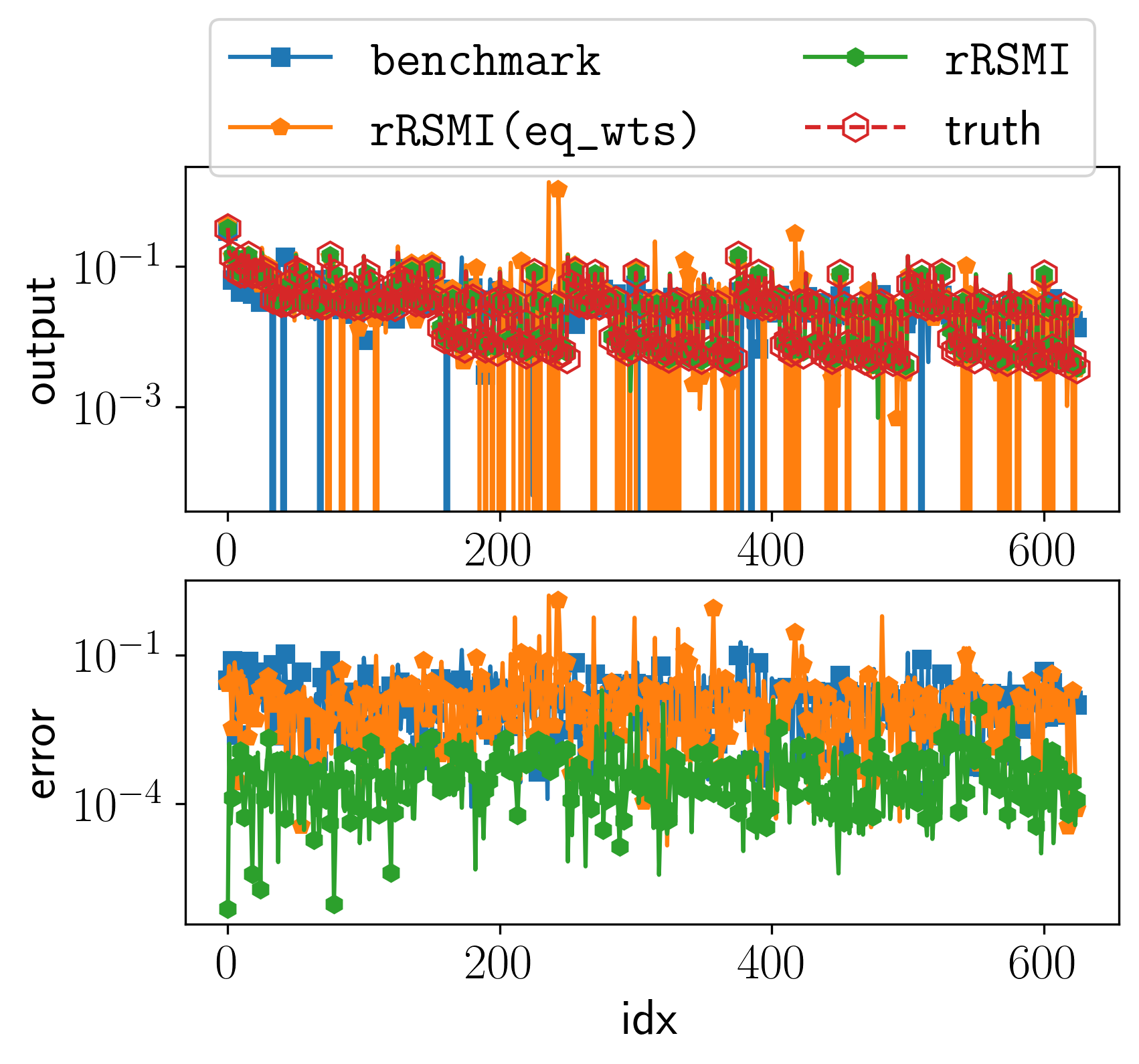}
		\caption{Output comparison.}
		\label{fig:thermal_symm_tfs}
	\end{subfigure}
	\caption{Thermal block example: Imposing symmetry leads to a faster decay of the singular values with \wnnreg, which results in a lower error in predictions on test data.} 
	\label{fig:thermal_symm_svd_tf}
\end{figure}

We now learn models with symmetric matrices $\bA_i, i\in\{0,\ldots,4\}$. The rest of the setup is the same as in the experiment without symmetry constraints. Analogous to previous experiments, imposing symmetry leads to a quicker decay of the singular values, and the model obtained with \wnnreg~has again the fastest decay. In contrast, the models obtained with \noreg~and \nnreg~are hardly compressible and prone to over-fitting to the training data. 
We construct models of order $r = 10$ via projection onto the subspace spanned by the first $r$ left-singular vectors and report the error of the models on the test data in \Cref{fig:thermal_symm_svd_tf,fig:thermal_med_err}. The model obtained with our approach \wnnreg~achieves the lowest median error by more than one order of magnitude. 

\subsubsection{Comparison with deep neural networks} 
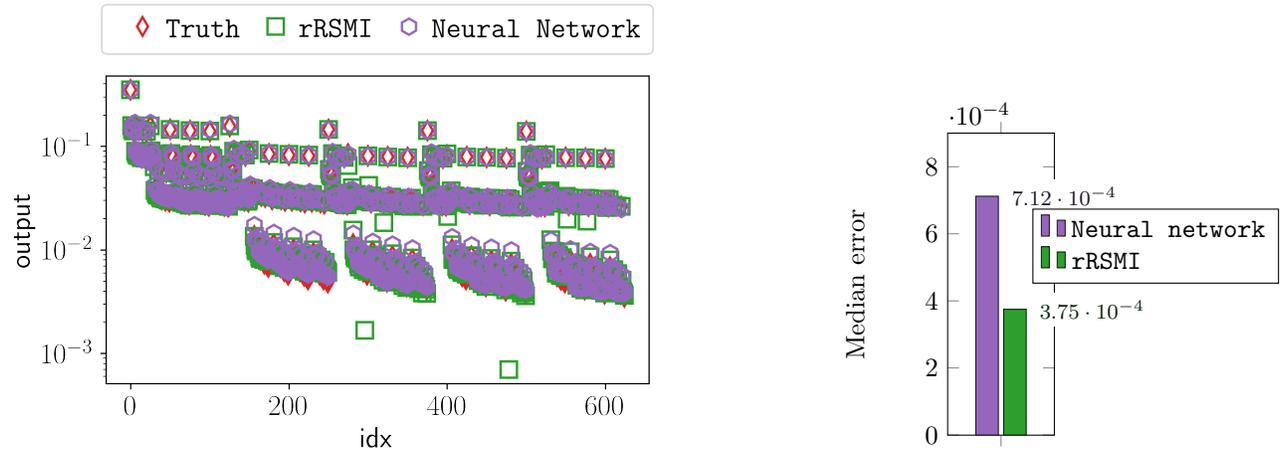
\begin{figure}[!tb]
	\includegraphics[width = 0.52\textwidth, trim = 0cm 0cm 0cm 0cm, clip]{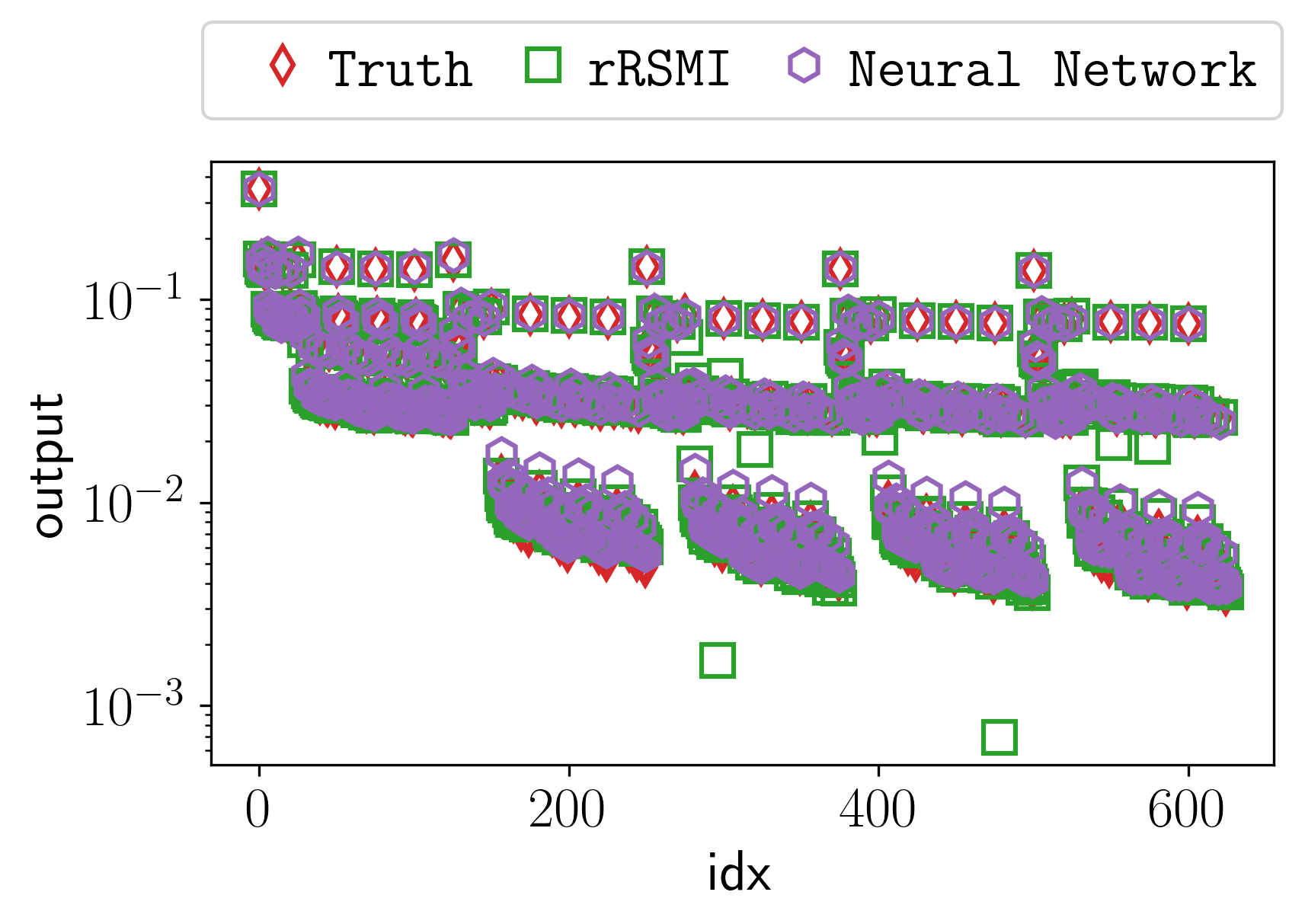}
	\hfill 
	\begin{tikzpicture}
		\begin{axis}[
			height=5.6cm,
			width=3cm,
			% ybar=0.5pt, % change this of you have more than one bar per column
			%x=3cm, % just play with the relation of these
			bar width=0.3cm, % just play with with the relation of these
			enlarge y limits={abs=0.001cm},
			% bar shift=2,
			ybar, 
			ymin=0, ymax = 0.0009,
			ylabel={Median error},
			% xmode=log,
			% symbolic y coords={%
			%     {approch1},
			%     {approch2},
			%     },
			xtick=data,
			xticklabels = {},
			x dir=reverse,
			nodes near coords={~\colorbox{white}{\pgfmathprintnumber{\pgfplotspointmeta}}},
			nodes near coords align={horizontal},
			nodes near coords style={font=\footnotesize},
			%  log origin=infty,
			point meta=rawy,
			legend style={at={(1.95,0.75)}, anchor=north, legend columns=1, /tikz/every even column/.append style={column sep=0.5cm}},
			cells={align=left},
			]
			
			\addplot[matplotlibcolor5!20!black,fill=matplotlibcolor5] plot coordinates {( 1.0,7.12e-4)};
			\addplot[matplotlibcolor3!20!black,fill=matplotlibcolor3] plot coordinates {(1.0,3.75e-4)};
			% \addplot plot coordinates {(1, 0.9)};
			
			\legend {\texttt{Neural network}, \wnnreg\qquad\qquad~~};
		\end{axis}
	\end{tikzpicture}
	\caption{Thermal block example: Plot (a) shows the outputs predicted by our approach \wnnreg~and by a neural network and compares them to the truth. Plot (b) compares the median error on the test data, where our approach achieves a factor two improvement over the network in this example.} 
	\label{fig:thermal_symm_NNs}
\end{figure}

We compare our \wnnreg~approach to a machine-learning method that fits a deep neural network to the input-output data. The machine-learning approach is a black box because it ignores the structure. The network architecture is fully connected. To train the network, we use the same training, validation, and testing data sets as for our \wnnreg~approach. To choose the hyper-parameters, in particular the number of hidden nodes and layers, we perform a grid search in $[4,8,16,32] \times [1,2,3]$, where $[4,8,16,32]$ and $[1,2,3]$ are possible numbers of nodes and layers, respectively. 
We then select the hyper-parameters that lead to the lowest error on the validation data set. In our case, we use one hidden layer and eight nodes.  
\Cref{fig:thermal_symm_NNs} shows the error of the trained network on the test data and compares it to the error obtained with the model of order $r = 10$ with our \wnnreg~approach with symmetry imposed. 
Our approach leads to a model that has a lower median error by a factor of two. 

\section{Conclusions}\label{sec:conclusions}
The proposed RSMI approach has two key properties that make it stand out: first, it leverages physical insights that are often given in the form of structure, such as symmetries, time delays, and degrees of time derivatives. The proposed RSMI approach bakes in such physical insights by imposing the corresponding structure onto the model. Critically, the structure is encoded in the model so that it cannot be violated rather than being weakly enforced via penalization terms. Additionally, the learned structured models can be realized as systems that describe the dynamics of states in the time domain and how they behave under control inputs, which is in contrast to black-box models that only match the input-output behavior. The second key property of RSMI is that the order of the inferred model, i.e., the number of degrees of freedom, is a training variable that is minimized during the optimization. The order of the model, therefore, is determined during training and does not have to be fixed based on heuristics a priori, as is typical in many other data-driven modeling approaches. By minimizing the order, RSMI learns models that are parsimonious because redundant degrees of freedom are eliminated when there are not sufficiently many training samples available to fix them. The numerical experiments demonstrate that the combination of structure preservation and optimizing the model order leads to models with low degrees of freedom while achieving high prediction capabilities. More broadly speaking, the results indicate that optimizing over the model architecture during the training can lead to more accurate predictions while keeping the number of degrees of freedom of the learned models low. There is a range of future research directions. For example, we have used (weighted-) nuclear norm minimization schemes as a proxy for the rank-minimization problems. In the future, we would like to explore other relaxations and computationally efficient approaches to obtain approximate solutions to rank-minimization problems.

\section*{Acknowledgements} B.P.~was supported by the National Science Foundation under Grant No.~2012250 and by the US Department of Energy, Office of Scientific Computing Research, Applied Mathematics Program (Program Manager Dr.~Steven Lee), DOE Award DESC0019334.

\bibliographystyle{siamplain}
\bibliography{mor}

\end{document}